\documentclass[5p,times]{elsarticle}

\usepackage{amssymb}
\usepackage{amsmath}
\usepackage{array}
\usepackage[caption=false,font=footnotesize]{subfig}
\usepackage{fixltx2e}
\usepackage{siunitx}
\usepackage{threeparttable}

\usepackage{hyperref}
\usepackage[utf8]{inputenc}
\usepackage{gensymb}
\usepackage{multirow}

\usepackage{xcolor}
\usepackage{soul}
\usepackage[short]{optidef}

\journal{Mechatronics}

\begin{document}

\begin{frontmatter}
\title{Reconfigurable Hydrostatics:\\Toward Versatile and Efficient Load-Bearing Robotics\tnoteref{t1}}

\tnotetext[t1]{Published version: Mechatronics (Elsevier). DOI:
\href{https://doi.org/10.1016/j.mechatronics.2025.103420}{10.1016/j.mechatronics.2025.103420}.}

\author{
  Jeff Denis$^a$}
\author{Frédéric Laberge$^a$}  
\author{Jean-Sébastien Plante$^a$} 
\author{Alexandre Girard$^a$} 
\affiliation{organization={$^a$Université de Sherbrooke},
            addressline={2500 Bd de l'Université}, 
            city={Sherbrooke},
            postcode={J1N3C6}, 
            state={Québec},
            country={Canada}}



\begin{abstract}

Wearable and legged robot designers face multiple challenges when choosing actuation. Traditional fully actuated designs using electric motors are multifunctional but oversized and inefficient for bearing conservative loads and for being backdrivable. Alternatively, quasi-passive and underactuated designs reduce the amount of motorization and energy storage, but are often designed for specific tasks.
Designers of versatile and stronger wearable robots will face these challenges unless future actuators become very torque-dense, backdrivable and efficient. 

This paper explores a design paradigm for addressing this issue: reconfigurable hydrostatics. We show that a hydrostatic actuator can integrate a passive force mechanism and a sharing mechanism in the fluid domain and still be multifunctional.
First, an analytical study compares the effect of these two mechanisms on the motorization requirements in the context of a load-bearing exoskeleton.
Then, the hydrostatic concept integrating these two mechanisms using hydraulic components is presented. A case study analysis shows the mass/efficiency/inertia benefits of the concept over a fully actuated one.
Then, experiments are conducted on robotic legs to demonstrate that the actuator concept can meet the expected performance in terms of force tracking, versatility, and efficiency under controlled conditions. The proof-of-concept can track the vertical ground reaction force (GRF) profiles of walking, running, squatting, and jumping, and the energy consumption is 4.8x lower for walking. The transient force behaviors due to switching from one leg to the other are also analyzed along with some mitigation to improve them.
\end{abstract}




\end{frontmatter}

\section{Introduction}
Mobile robots that must bear their own weight have conflicting design requirements. For reasonable autonomy, their actuators should be lightweight and efficient, but at the same time they need good backdrivability for good physical interaction with their environment, e.g., with the ground for a legged robot or with the user for an exoskeleton; and still to be useful in various situations (multifunctional), they should have high strength and power levels.

\begin{figure}[t!]
\centering
\includegraphics[width=0.99\columnwidth]{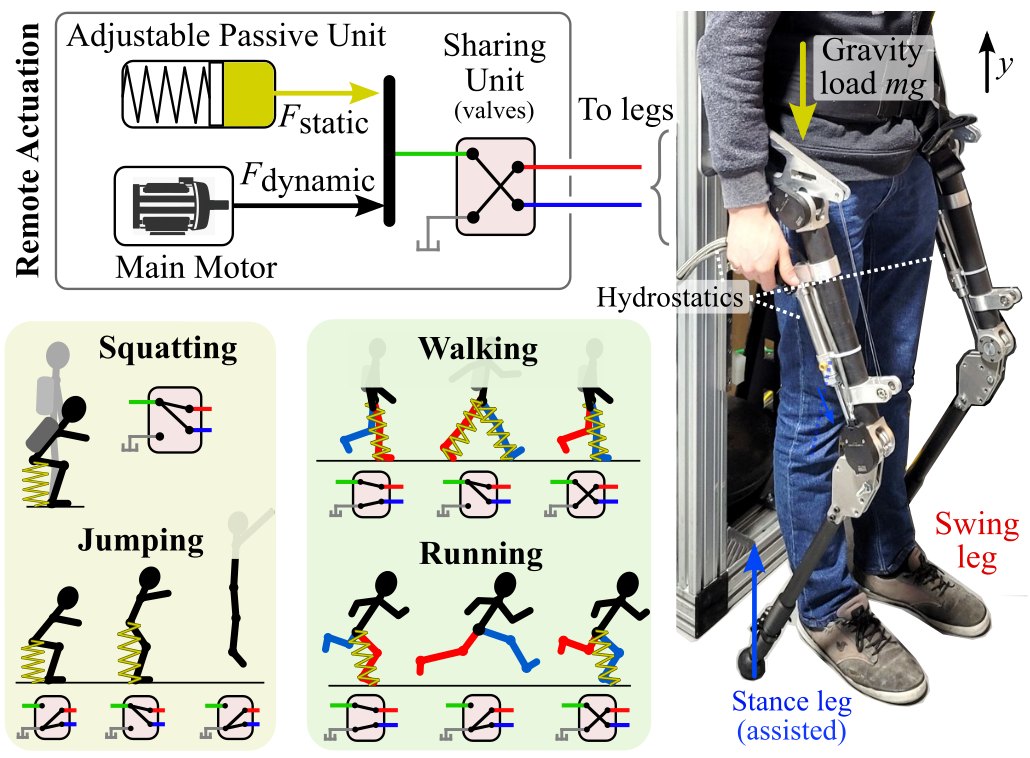}
\caption{Overview of the proposed multifunctional actuator. The same color coding is used through the paper for their corresponding variables and signals.}
\label{fig:overview}
\end{figure}

The recent research in legged robots and exoskeletons push mainly for lightly geared electric motors for their high velocity and relatively good torque density and backdrivability \cite{seok_design_2015,saloutos_design_2023,elery_design_2020}.
These actuators 1) have low inertia for better interactions and simpler force control, 2) have high transmission efficiency, and 3) enable good energy regeneration at batteries \cite{seok_design_2015,kashiri_overview_2018}. However, by producing high electromagnetic torque, their motors heat significantly. For instance, the running legged robot Cheetah could regenerate most of its negative power to the battery, but 74\% of its energy consumption was due to motor heating \cite{seok_design_2015}.
Over lightly geared motors, for even higher backdrivability, continuous slippage clutches like magnetorheological (MR) clutches were also proposed \cite{khazoom_design_2019}.
All these single-actuator-per-DOF (degree of freedom) strategies race for the highest force and power density actuator without compromising versatility.

To relax the motorization requirements of each DOF and improve efficiency, a passive element like a spring can be added, as in series-elastic and parallel-elastic actuators (SEAs, PEAs) \cite{grimmer_comparison_2012}. Some exoskeletons are even passive only \cite{elliott_design_2014} \cite{dollar_design_2008} and do not need active motorization. However, these concepts are typically working for a specific task only.

To reduce the number of motors in multiple DOFs systems, actuation may also be shared. For instance, a hydraulic pump can distribute its power to many DOFs via servovalves and achieve highly dynamic and powerful tasks when this power is concentrated on a few joints, as in the jumping motions of the humanoid Atlas from Boston Dynamics \cite{gizzo_boston_2016}. This multifunctionality, however, causes energy losses at idle joints and requires a heavier centralized motor and storage. Pneumatic power can likewise be shared across multiple joints via solenoid valves and soft actuators \cite{zhou_portable_2024}, but depressurizing large volumes of compressible gas when assistance is removed leads to significant energy loss.
An alternate way to share actuation is underactuation, for instance exoskeletons that can fully disconnect the actuator from an unused joint to power another one, like for assisting the left and the right legs during stance \cite{asbeck_multi-joint_2015}. However, so far, these concepts can only assist specific tasks. 

As discussed above, quasi-passive actuators and sharing actuation were proposed as lighter/more efficient/cost effective solutions compared to fully actuated devices, but at the expense of low versatility.
In a previous paper, the authors proposed to leverage hydrostatic transmissions to implement various hybrid actuation modes such as gravity-load compensation \cite{denis_multimodal_2022}. This paper describes, analyzes and tests a new reconfigurable hydrostatic topology that is multifunctional and combines both 1) an adjustable static force compensation, and 2) sharing the same actuation through multiple degrees of freedom. An overview is given in Fig.~\ref{fig:overview}.
%
%
The first contribution is a hydrostatic actuator topology that combines both two mechanical principles and still assist various tasks. The second contribution is an analysis of the effect of these units on the motorization requirements for different tasks and how the total actuation mass, efficiency and inertia compares for a given case study. The third contribution is a preliminary experimental study evaluating the performance of the proposed actuator on load-bearing robotic legs following predefined force profiles for walking, running, squatting, and jumping scenarios, along with strategies to improve switching using valves.



%
Section~\ref{section:related_work} goes deeper into the state-of-the-art regarding passive load balancing and actuator sharing through multiple joints.
Section~\ref{section:mass_efficiency_analysis} analyzes the effect of these two principles on the motorization requirements.
Section~\ref{section:proposedDesign_and_CaseStudy} presents the proposed hydrostatic actuator and a case study analysis to compare the total mass, efficiency and inertia advantage with off-the-shelf components.
Section~\ref{section:proof-of-concept} presents a proof-of-concept and a basic controller which is tested in section~\ref{section:experiments} to validate experimentally that it can track the force profiles of walking, running, squatting and jumping and be more efficient as well.
Finally, section~\ref{section:switching_mitigations} presents control mitigation using the valves to improve the switching effects.

\section{Related Work}
\label{section:related_work}
This section presents how the state-of-the-art locomotive robots and lower-limb exoskeletons applied static load compensation and actuator sharing, these two operating principles being integrated into the actuator of this paper. Although being promising for efficiently bearing payloads and for reducing the number of motors, most existing designs lack versatility.


\subsection{Efficient Static Load Compensation}
\begin{figure}[h!]
\centering
\includegraphics[width=0.999\columnwidth]{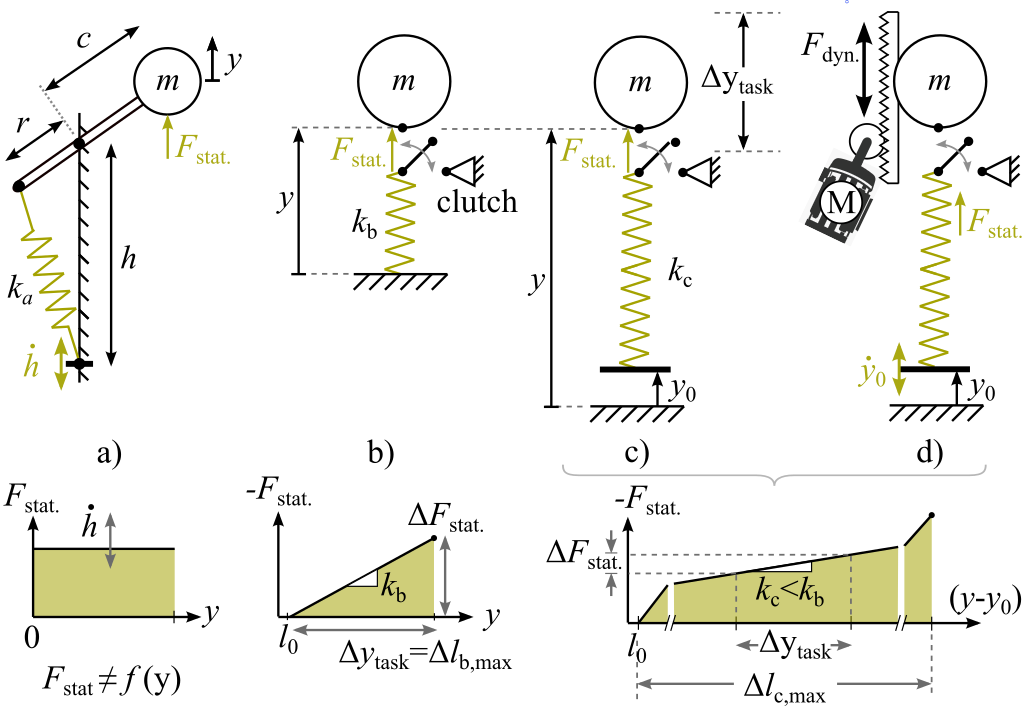}
\caption{Spring-based mechanisms in evolution of complexity and functionalities: a) clutchable spring, b) static balancing mechanism with adjustable attachment, c) a + large spring + adjustable $y_0$, d) c + parallel actuator. $\Delta y_\text{task}$ is the vertical travel needed for a given task and $l_0$ is the spring length at rest.}
\label{fig:gravity_compensation_concepts}
\end{figure}

Legged locomotion is an energy recycling process. For conservative force problems like walking/running at constant speed on flat ground, passive designs are likely more efficient than active designs using motors, but not as versatile. Fig.~\ref{fig:gravity_compensation_concepts} shows how increasing the complexity of passive mechanisms can lead to more functionalities.
First, a parallel grounded spring could bear a load without needing any energy, just like a car's suspension do.
Quasi-passive exoskeletons often use this principle: a parallel spring fits the natural force-displacement behavior of a joint for the stance phase, as long as a clutching mechanism can disconnect the swinging leg \cite{elliott_design_2014} \cite{dollar_design_2008}. This approach shown by Fig.~\ref{fig:gravity_compensation_concepts}b is, however, typically task-specific since the ideal joint stiffness varies widely between tasks like squatting, walking and running.

For more versatility in passive actuation, static load compensation mechanisms are alternatives to matching a joint's stiffness. Such mechanism generates a static force $F_\text{stat.}\neq f(y)$ that is independent of the output motion $y$. This way, one cancels out the gravity force acting on the beared mass $m$ whenever task is done. First, counterweights can balance static loads \cite{lacasse_design_2013}, but it increases the total beared weight and inertia. A second approach is the use of a spring and a non-linear mechanism (see Fig.~\ref{fig:gravity_compensation_concepts}a), as done in \cite{banala_gravity-balancing_2006} for balancing a human leg or in \cite{zhang_novel_2024} for a robotic knee using a nitrogen gas spring and a non-linear cam mechanism. Finally, static load compensation may be approximated with a concept like Fig.~\ref{fig:gravity_compensation_concepts}b if the spring is large and compliant enough, like in Fig.~\ref{fig:gravity_compensation_concepts}c. Indeed, when the maximum spring elongation is much higher than the output travel for a given task, $\Delta l_\text{max} \gg \Delta y_\text{task}$, $F_\text{stat.}$ may not vary much. A spring generating the same maximum force but with an elongation $n$ times higher, $\Delta l_\text{max,c}=n\Delta l_\text{max,b}$, is also $n$ times more compliant, $k_\text{c}=k_\text{b}/n$, leading to less spring force variation over $\Delta y_\text{task}$. The downside is that the spring is also potentially $n$ times heavier\footnote{\label{footnote:massSpring}If we assume that the mass of a spring is proportional to its energy stored. Indeed, the potential elastic energy capacities for Fig.~\ref{fig:gravity_compensation_concepts}b and \ref{fig:gravity_compensation_concepts}c are $E_\text{b}=k_\text{b}{\Delta l_\text{b,max}}^2/2$ and $E_\text{c}=k_\text{c}{\Delta l_\text{c,max}}^2/2=nE_\text{b}$.}. 

The above static compensation concepts lack the flexibility of adjusting the passive force $F_\text{stat.}$ on purpose. For the concept of Fig.~\ref{fig:gravity_compensation_concepts}a, this is possible by moving the attachment points of the spring since $F_\text{stat.} = k_{a}hr/c$ \cite{barents_spring--spring_2011}. This was done by \cite{kim_variable_2023} for an efficient robotic leg using linear rails and a low-power non-backdrivable actuator along the leg, but this implementation is rather complex. For the concept of Fig.~\ref{fig:gravity_compensation_concepts}c, the force is $F_\text{stat.}=-k(y-l_0-y_0)$. For adjustable passive force, one could change $k$ in real-time just like variable stiffness actuators do. Otherwise, one can instead move the spring attachment point $y_0$ with a low-power motor. The feasibility of this solution depends on how strong and how compliant the spring has to be, how fast ($\Delta t$) the passive assistance shall be adjusted and how energy-dense and power-dense the spring and geared motors are, respectively $^{\ref{footnote:massSpring},}$ \footnote{The mean power of the adjusting motor is given by $E_\text{c}/{\Delta t}$, i.e. how fast the motor has to fully compress the spring. More powerful gearmotors are heavier.}.

So far, concepts 2a--c rely solely on a passive actuator force but for many applications, this is not sufficient. Indeed, dynamic force capabilities $F_\text{dyn.}$ are needed to move the legs of a legged robot and can benefit to exoskeletons to follow the correct GRF trajectories of different tasks and improve the transitions between the stance and the swing phases. In Fig.~\ref{fig:gravity_compensation_concepts}d, a parallel active actuator is added to the static load compensation system. This results in a parallel-elastic actuator but with a very compliant spring and an adjusting mechanism. A research robotic leg implements this principle in \cite{tsagarakis_asymmetric_2013,roozing_development_2016} with a high-power series-elastic actuator that is coupled in parallel to a passive unit consisting of an elastic band being stretched by a non-backdrivable screw to balance a payload. A clutch can also actively disconnect the elastic band for the swing phase and for charging powerful jumps. The authors present a torque distribution control strategy and measured a 65\% energy consumption reduction for squatting motions when using the passive unit. The concept was pushed further for monoarticular and biarticular 3-DOF leg designs in \cite{roozing_design_2018} Recently, Fan et al. introduced a highly efficient legged robot capable of supporting variable payloads during the stance phase by simply blocking a hydrostatic circuit—a strategy that is viable for walking, provided the body height remains constant \cite{fan_load-carrying_2024}. However, this approach is unsuitable for exoskeletons, as it would resemble walking on rigid canes, and it cannot accommodate tasks involving vertical motion such as running, squatting, or stair climbing. In a related study \cite{fan_hyexo_2024}, the same authors proposed a passive load-bearing exoskeleton that uses a shared hydraulic accumulator. In this design, the assistive pressure can be modulated by precisely timing the opening of valves.

%
%
%

To date, gravity compensation in robotics is typically restricted to specific tasks. Compensating for heavy payloads—such as those in human-scale robots—can be impractical, particularly when these springs are collocated along the legs. Instead, relocating the springs and employing high energy-density ones could significantly enhance the effectiveness of this approach.

\subsection{Actuator Sharing Through Multiple Joints}
Most locomotion and wearable robots have one actuator per DOF. Fully actuated designs have controllability at all time, but can be overdesigned since torque requirements vary widely between tasks. For instance, during the stance phase, the human knee torque is relatively low for walking (0.5--0.7~\si{\newton\meter\per\kg}) \cite{fukuchi_public_2018} compared to the torque for running (3.0--4.0~\si{\newton\meter\per\kg}) \cite{fukuchi_public_2017}. Synergies in locomotion can be exploited to reduce the number of required actuators while keeping the possibility assisting many joints. The joints of the left and the right legs either work in-phase (e.g., squats, sit-to-stands and jumps), sharing the same force profiles, or out-of-phase (e.g., the stance phases of walking and running). 
%
%
Underactuation draw increasing attention in the recent years in lower-limb soft exosuits, with remote motorization in the back. A back exosuit in \cite{arens_preference-based_2025} uses a single motor–strap mechanism to generate an extensor moment at the hips and lumbar joints. In \cite{ko_waist-assistive_2018}, with a differential gearbox, one motor can assist both hips during lifting motions, without hindering out-of-phase motions like walking. In \cite{lanotte_design_2020}, a second differential gearbox is added to power the lower-back joint as well, so three joints for a single motor. In \cite{asbeck_multi-joint_2015, tricomi_underactuated_2022}, a single motor assists the stance phases of walking of both hips thanks to the buckling effect of cable transmissions which decouple the motor from the swinging leg. 
%
%
Other authors proposed multiarticular actuation like for the Myosuit which successfully reduces muscle efforts due to gravity loads by assisting simultaneously the knee and hip of one leg \cite{schmidt_myosuit_2017}.

All current exoskeletons sharing the same actuation for the left and right legs have the same limitation: they can either assist in-phase or out-of-phase tasks but never both (at best, they just do not hinder it). Moreover, all existing designs rely on cable transmissions or differential gear mechanisms only. A multifunctional underactuated exoskeleton that shares actuation for both legs is yet to be developed.

In this section, it was stated that a static force passive unit and a sharing unit can benefit to a locomotive robotics. In the next section, we calculate how these two functions can actually relax the total motorization requirements in the context of a lower-limb exoskeleton.

\section{Effect on the Motorization Requirements}
\label{section:mass_efficiency_analysis}
We discussed two mechanical principles that can potentially improve lightly geared actuator designs: 1) adding an adjustable static passive force unit and 2) adding a sharing unit. This section presents a simplified yet insightful analysis of the expected benefits of integrating these functions into a load-bearing exoskeleton design. It is found that when combining the two mechanical principles, the total RMS force to be generated by the motorization gets 2.7x to 7.8x lower.

\begin{figure}[h!]
\centering
\includegraphics[width=0.99\columnwidth]{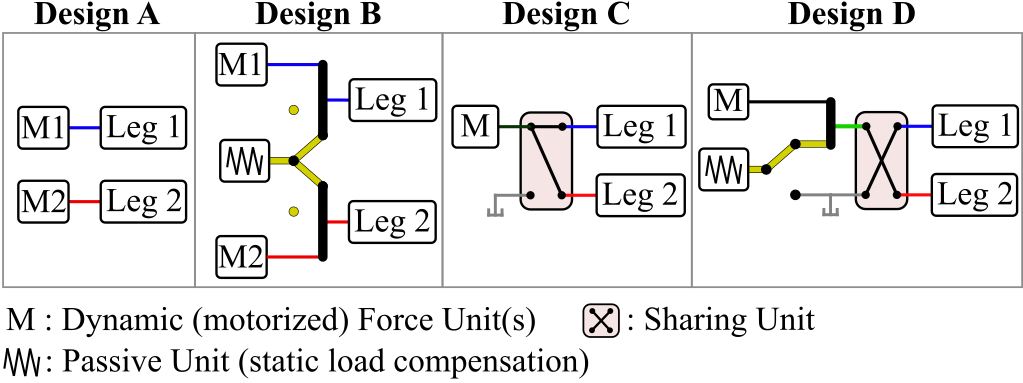}
\caption{The four generic designs compared in this analysis: A) fully actuated, B) with passive unit, C) with sharing unit, D) with both units.}
\label{fig:designsABCD}
\end{figure}

Let's introduce a 2-DOF multitask exoskeleton that assists the vertical ground reaction forces (GRFy) of the right leg $f_1(t)$ and the left leg $f_2(t)$, for instance to reduce the musculoskeletal stress and the metabolic expenditure when carrying payloads like in \cite{schmidt_myosuit_2017} \cite{mawashi_uprise_nodate}. The following analysis is valid regardless of how these functions are mechanically obtained. Four generic designs are described by Fig.~\ref{fig:designsABCD} and are compared in this section for four tasks: walking, running, jumping and sit-to-stands. It is worth noting that Design~B is also compatible with legged robotic systems, whereas C and D are not directly applicable in such contexts, as they require actuators to swing the legs.

%

\begin{figure*}[t!]
\centering
\includegraphics[width=0.99\textwidth]{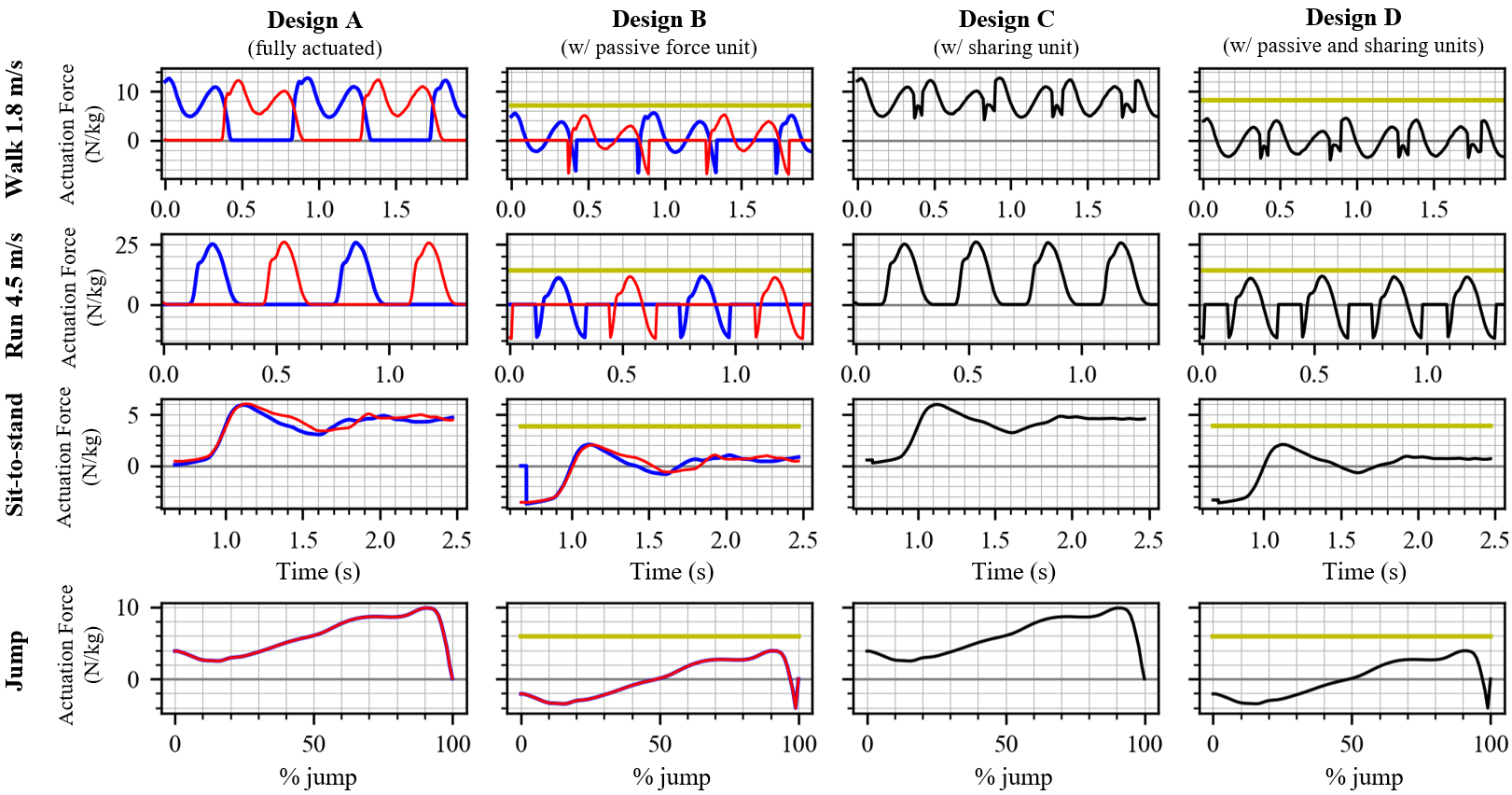}
\caption{Vertical GRF to be generated by the actuation for walking, running, sit-to-stands and jumping for the generic designs A--D (blue, red and black lines are the dynamic forces for the right, left or for both legs, respectively, and yellow line for the static force offset when applicable.).}
\label{fig:requirements_curves}
\end{figure*}

Fig.~\ref{fig:requirements_curves} (first column) gives the typical normalized vertical GRF curves used, in $\si{\newton\per\kilo\gram}$ or $\si{\meter\per\second\squared}$ of young human adults, obtained from papers and public datasets \cite{fukuchi_public_2018,fukuchi_public_2017,liang_asian-centric_2020, cormie_power-time_2008}.
%

\subsection{Comparison Metrics} Here we introduce metrics to compare the requirements for the dynamic force source of an exoskeleton system. For long cyclic tasks like walking and running, the force output of an electric motor is most likely limited by its continuous rated torque. To prevent overheating, it must exceed the root mean squared (RMS) torque required by the task. The RMS torque is also indicative of resistive (Joule) losses, since it is proportional to the winding current $I$, $\overline{P}_\text{Joule}=R\left( \overline{I(t)^2} \right)=R{I_\text{rms}}^2$.
Joule losses are often the dominant source of energy consumption lightly geared motors \cite{seok_design_2015}. Consequently, minimizing the RMS force generated by the dynamic actuator, $f_\text{dyn,rms}$, is essential for achieving a lightweight gearmotor and compact battery system.
Another relevant metric for evaluating energy efficiency in cyclic tasks is the mean absolute power, $\overline{\lvert P_\text{dyn}(t)\rvert}$, delivered by the actuator. In practice, mechanical power is lost during power generation (${P_\text{dyn}}^{+}(t)$) and regenerative braking (${P_\text{dyn}}^{-}(t)$). The power drawn from the battery to drive the dynamic force unit is expressed by equation~\ref{eq:power_storage} where $\eta_\text{gen.}$ and $\eta_\text{regen.}$ are the efficiency when generating and dissipating power, respectively:

\begin{equation}
    \label{eq:power_storage}
    P_\text{battery}(t)=P_\text{Joules}(t)+\frac{ {P_\text{dyn.}}^{+}(t) }{\eta_\text{gen.}}+\eta_\text{regen.}{P_\text{dyn.}}^{-}(t)
\end{equation}

In energy-conservative cyclic tasks, the average positive and negative powers tend to cancel out, $\overline{{P_\text{dyn.}}^{+}(t)}=-\overline{{P_\text{dyn.}}^{-}(t)}$ \footnote{If no energy lost between the dynamic force unit and the leg output.}. Under this assumption, the average battery power becomes:

\begin{equation}
    \label{eq:power_storage2}
    \overline{P}_\text{battery}(t)=\overline{P}_\text{Joules}(t)+\Big(\frac{1}{\eta_\text{gen.}}-\eta_\text{regen.}\Big)
    \frac{\overline{\lvert P_\text{dyn.}(t)\rvert } }{2}
\end{equation}
Equation~\ref{eq:power_storage2} highlights the portion of energy that cannot be recovered due to limited gearbox efficiency. For instance, with $\eta = 90\%$, only 21\% of the mechanical power is lost, whereas a less efficient, highly geared system with $\eta = 70\%$ results in a 73\% loss. Regardless of $\eta$, however, a lower value of $\overline{\lvert P_\text{dyn}(t)\rvert}$ directly translates into reduced energy consumption, as less mechanical power flows through the actuator.

%
Another important metric is the peak force $f_\text{dyn.peak}$, especially for low duty cycle tasks like jumping and sit-to-stands where the motors do not have the time to heat. The transmission sizing (e.g., gearbox) is also often based on this metric. Finally, for powerful tasks like jumping and running, the maximum speed $V_\text{dyn.}$ can also be limiting factor.

\subsection{Calculation}
Here we describe how the metrics are computed for Designs A--D and all the tasks. For Design A, the two motors should directly track the output force curves of each leg, meaning that for out-of-phase tasks (walking and running), each motor is used only 25-50\% of the gaits. The RMS dynamic force of each leg $i$ is:

\begin{equation}
    f_{\text{A,dyn,rms,}i} =\sqrt{ \frac{1}{T} \int_0^T  {f_i(t)}^2 \mathrm{d}t}
\end{equation}

For Design B, the passive unit offsets by a static force $f_\text{stat.}$ the dynamic force of each motor. We assume this passive force constant\footnote{Perfect static load balancing is feasible with a non-linear transmission (see Fig.~\ref{fig:gravity_compensation_concepts}a) or approximated with a very compliant spring (see Fig.~\ref{fig:gravity_compensation_concepts}c).}.
%
%
The static force offset that minimizes $f_\text{dyn,rms}$ is given by solving numerically equation~\ref{eq:optimization}:

\begin{equation}
    \begin{split}
        {f_{\text{B,stat,}i}}^*~=~&\operatorname*{argmin}_{f_{\text{B,stat,}i}} {~~\underbrace{\sqrt{ \frac{1}{T} \int_0^T  {  f_{\text{B,dyn,}i}(t)  }^2 \mathrm{d}t}}_{f_{\text{B,dyn,rms,}i}}} \\
        \text{subject to~~~~}& f_{\text{B,dyn,rms},i} \leq \frac{\text{max} |f_{\text{B,dyn,}i}(t)| }{3}
    \end{split}
    \label{eq:optimization}
\end{equation}


where:
\begin{equation}
    f_{\text{B,dyn,}i}(t)= 
    \left\{\begin{matrix}
    f_i(t)-f_{\text{B,stat,}i } & \text{if } f_i(t) > 0 \\    
    0 & \text{otherwise\footnotemark }
\end{matrix}\right.  
\label{eq:f_dynB}
\end{equation}
\addtocounter{footnote}{-2}
\footnotetext{In the aerial phase ($f_i (t)=0$), the static force unit is disconnected.}%
The constraint limits the peak force needed at the motors to three times its rated force as for typical electric motors \cite{robodrive_motor_2021}.

For Design~C, a single dynamic force unit is designed to track the total force $f_\text{tot}$ of both legs:

\begin{equation}
    f_\text{tot}(t)=f_\text{1}(t)+f_\text{2}(t)
    \label{eq:output_force}
\end{equation}

In a differential-like transmission mechanism where both legs are connected to a single actuator, the actuator's speed $V_\text{sharing}$ and force $f_\text{sharing}$ are:

\begin{equation}
    V_\text{sharing}(t) = V_1(t) + V_2(t)
\end{equation}
\begin{equation}
    f_\text{sharing}(t) = f_1(t) = f_2(t)
    \label{eq:differential_force}
\end{equation}

From equations \ref{eq:output_force} and \ref{eq:differential_force}, when the two legs are connected to the shared actuator ($N_\text{legs}$=2), the available GRF is doubled. For a given task, the required force at the motor is then:

\begin{equation}
    f_\text{C,dyn.}(t) = 
    \left\{\begin{matrix}
    \frac{ f_\text{tot}(t)}{ N_\text{legs}(t) }, & \text{if~} N_\text{legs}(t) > 0 \\    
    0, & \text{otherwise}
\end{matrix}\right. 
\label{eq:C_motorForce}
\end{equation}

Finally, for Design D, the static force is found by solving equations~\ref{eq:optimization} and \ref{eq:f_dynB}, but replacing $f_{i}(t)$ by equation~\ref{eq:C_motorForce}.

\begin{table}[h!]
\footnotesize
\renewcommand{\arraystretch}{1.0}
\caption{Performance metrics generated by the motor(s) (M) when leg(s) are on ground for all tasks and designs}
\label{table:requirements_RMS_peak}
\centering
\begin{tabular}{| c | c | c c | c c  | c  | c |}

\hline 
 \multicolumn{2}{|r|}{Design} &  \multicolumn{2}{c|}{\textbf{A}} & \multicolumn{2}{c|}{\textbf{B}}  & \textbf{C} & \textbf{D} \\
\cline{3-8}
  \multicolumn{2}{|r|}{} & \multicolumn{2}{c|}{-}  & \multicolumn{2}{c|}{Passive} & - & Passive \\
\multicolumn{2}{|r|}{}  & \multicolumn{2}{c|}{-} & \multicolumn{2}{c|}{-} & Sharing & Sharing \\
\hline
 \multicolumn{1}{|l|}{Task} & Metric* & $\text{M}_1$ & $\text{M}_2$ & $\text{M}_1$ & $\text{M}_2$ & M & M  \\
\hline 
\hline

 & $\boldsymbol{f}_\textbf{dyn,rms}$ & \textbf{6.5} & \textbf{6.5} & \textbf{2.3} & \textbf{2.3} & \textbf{8.6} & \textbf{2.4} \\
 Walk & $f_\text{dyn,peak}$ & 13.4 & 13.4 & 7.0 & 7.0 & 13.7 & 5.5 \\
 \scriptsize(1.8 m/s)  & $\overline{\lvert P_\text{dyn.}\rvert }/2$  & \textcolor{black}{0.47} & \textcolor{black}{0.47} & \textcolor{black}{0.14} & \textcolor{black}{0.14} & \textcolor{black}{0.93} & \textcolor{black}{0.22} \\
 & $V_\text{dyn.}$ & 0.4 & 0.4 & 0.4  & 0.4 & 0.7 & 0.7 \\
\hline 
  & $\boldsymbol{f}_\textbf{dyn,rms}$ &\textbf{9.7} & \textbf{9.7} & \textbf{5.1} & \textbf{5.1} & \textbf{13.7} & \textbf{7.2} \\
 Run &$f_\text{dyn,peak}$  & 26.6 & 26.6 & 13.9 & 13.9 & 27.4 & 14.1 \\
  \scriptsize(4.5 m/s) & $\overline{\lvert P_\text{dyn.}\rvert }/2$  & \textcolor{black}{1.0} & \textcolor{black}{1.0} & \textcolor{black}{0.72} & \textcolor{black}{0.72} & \textcolor{black}{2.0} & \textcolor{black}{1.4} \\
 &$V_\text{dyn.}$ & 1.0 & 1.0 & 1.0 & 1.0 & 1.0 & 1.0 \\
\hline 

Jump  &$f_\text{dyn,peak}$  & 9.9 & 9.9 & 4.1 & 4.1 & 9.9 & 4.1 \\
\scriptsize(0.4 m) &$V_\text{dyn.}$ & 2.7 & 2.7 & 2.7 & 2.7 & 5.4 & 5.4 \\
\hline 

 Sit-to- &$f_\text{dyn,peak}$ & 6.0 & 6.0 & 3.7 & 3.7 & 6.0 & 3.6 \\
 stand &$V_\text{dyn.}$ & 0.7 & 0.7 & 0.7 & 0.7 & 1.4 & 1.4 \\
\hline
\end{tabular}
\begin{tablenotes}
      \footnotesize
      \item *All metrics are normalized by the user's bodyweight, $f_\text{dyn,rms}$ and $f_\text{dyn,peak}$ in \si{\newton\per\kg}, $\overline{\lvert P_\text{dyn.}\rvert}$ in \si{\watt\per\kg}, with the exception of $V_\text{dyn.}$ in \si{\meter\per\second}.
    \end{tablenotes}
\end{table}

\subsection{Results and discussion}
Fig.~\ref{fig:requirements_curves} shows for each task the analytical force trajectories required at the motor(s) for Designs A--D, and the static force offset computed. Table~\ref{table:requirements_RMS_peak}summarizes the key performance metrics for each design (A–D) and task. These values reflect the requirements imposed on the main dynamic actuator(s) and allow simplified yet meaningful comparisons between the designs. 

The results first reflect the importance of thermal considerations over peak force capability for walking and running, assuming a motor can generate up to three times its nominal torque. Peak force is more relevant for short-duration tasks like jumping or sit-to-stand.

The implementation of a passive force unit in \textbf{Design B} significantly reduces the demands on the dynamic force actuator. For instance, the RMS force is reduced by up to 2.8× in walking and 1.8× in running. Assuming the same gearmotor for Designs A and B, this implies that Design~B could either support a 2.8× heavier payloads without overheating, or generates down to 7.8× ($2.8^2$) less motor heat for the same payload. Moreover, the transmission losses are also reduced due to less power flow in the dynamic force unit ($\overline{\lvert P_\text{dyn}(t) \rvert }$).

Regarding \textbf{Design~C}, the advantages of actuator sharing are less straightforward. While the number of gearmotors is effectively halved, the shared actuator must deliver twice the speed for combined-leg tasks such as sit-to-stand and jumping, and slightly higher RMS force during out-of-phase tasks like walking. In turns, the comparison necessarily involves different gearmotor designs\footnote{Gearmotor sizing depends on the torque, speed and mechanical transparency desired.}, making a direct comparison with Design~A inconclusive. Table~\ref{table:requirements_RMS_peak} remains a valuable input for the design of such systems, as in the case study of Section~\ref{section:proposedDesign_and_CaseStudy}.

When both principles are combined in \textbf{Design D}, the results of Design~B et and C combines too. The solution appears superior to Design~A, particularly for tasks dominated by conservative loads and for low-speed requirements when both legs are assisted.

This analysis showed how sharing and static load compensation impact the motorization requirements of a load-bearing exoskeleton. However, implementing these functions is not trivial and involves others components. The next section presents a novel design that implements for the first time both principles into a single device. It also presents a case study to better compare Designs~A and D in a more detailed scenario.


\section{Reconfigurable Hydrostatics Proposition and Case Study}
\label{section:proposedDesign_and_CaseStudy}
So far this paper discussed the potential of sharing actuation and adjustable static load compensation for locomotive robots and exoskeletons. Here we propose a design opportunity to leverage hydrostatic transmissions and hydraulic components to implement both functions. Then, the results from section~\ref{section:mass_efficiency_analysis} are used in a specific case study analysis showing that the new design can be more backdrivable and consumes 3.9x less power for the same performance and total actuation weight.

\subsection{Proposed Design}
\begin{figure}[h!]
\centering
\includegraphics[width=1\columnwidth]{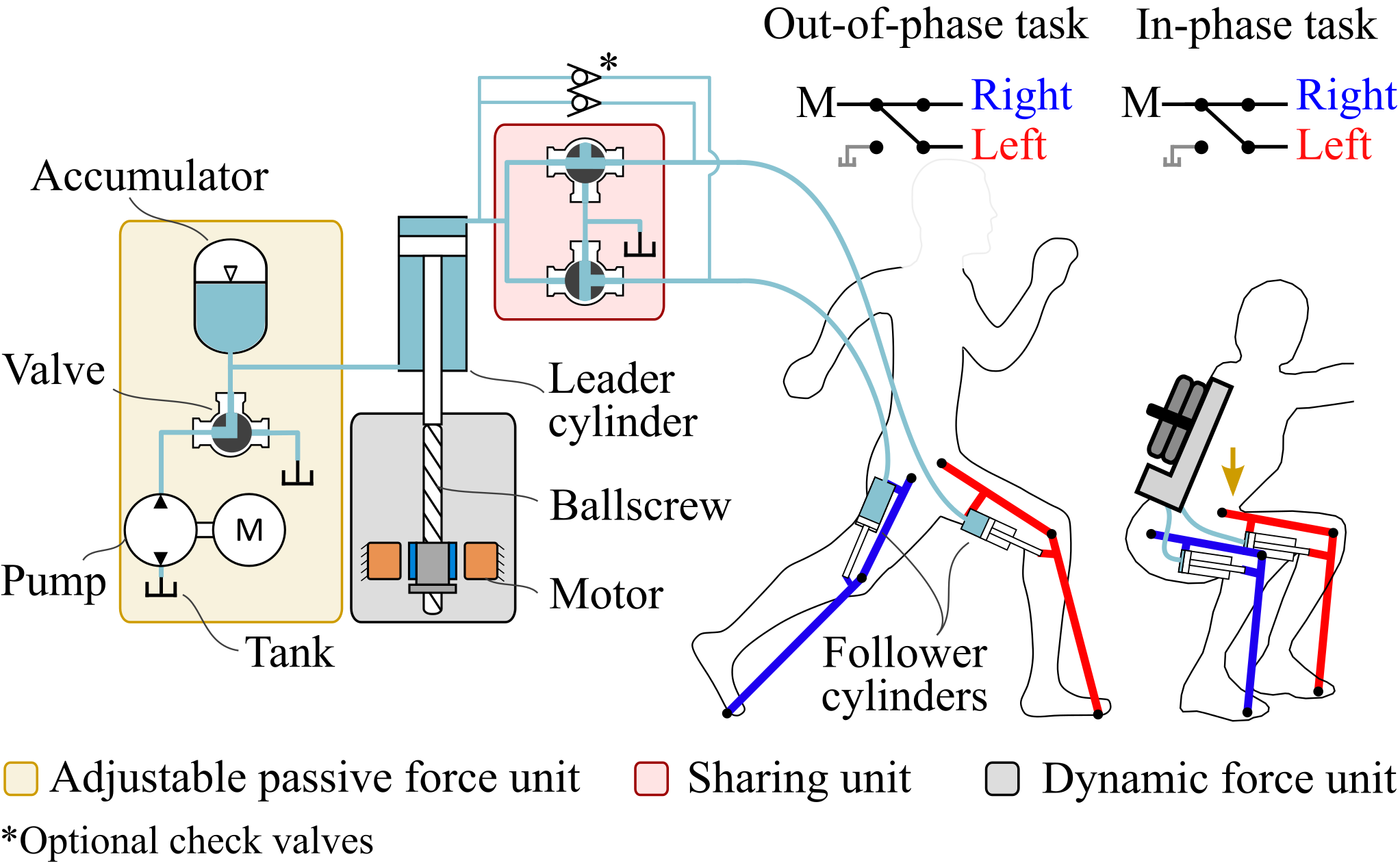}%
\caption{Actuator proposition combining actuation sharing and an adjustable passive force unit into single topology for a lower-limb exoskeleton.}
\label{fig:proposed_concepts}
\end{figure}

Hydrostatic transmissions are an alternative to cable transmissions to remotely actuate leg joints, leading to low-inertia robotic legs/exoskeletons. As discussed in section~\ref{section:related_work}, adjustable static load compensation was proposed previously for a robotic leg using elastic bands, a screw transmission and a clutch. The proposed design in Fig.~\ref{fig:proposed_concepts} includes an equivalent adjustable passive force unit (yellow) relying instead on a pump, an accumulator and a motorized valve.
Since it acts in parallel, the passive force unit does not impede the force bandwidth of the main motor and may not increase significantly the backdriving force if designed properly.

The proposed design in 
Fig.~\ref{fig:proposed_concepts} also includes a sharing unit (red) that uses an array of valves to share actuation for both legs. This design can assist both in-phase and out-of-phase tasks thanks to the modularity of hydraulic valves. Note that the state-of-the-art exoskeletons rely on buckling cables or differential gearboxes to either assist in-phase or out-of-phase tasks, as discussed in section~\ref{section:related_work}. To assist both type of tasks, these state-of-the-art concepts would need extra switching mechanisms like clutches. In Fig.~\ref{fig:proposed_concepts}, during the swing phase, the user backdrives the swinging leg being connected to the low-pressure tank. For in-phase tasks like squatting, the pressure is shared and the flow is split between the legs. The hydrostatic transmissions is thus a fluidic differential.

To the authors’ knowledge, this is the first robotic design to integrate both an adjustable passive force unit and a sharing unit. This design is implemented experimentally in this work. The authors believe that merging these two design principles would be more complex using clutches, brakes and gearboxes, instead of hydraulic components like here. Still, the passive unit and sharing functions needs extra components. The following case study compares this concept to a fully actuated one by the selection of off-the-shelve components.

\subsection{Case Study Comparison}
Section~\ref{section:mass_efficiency_analysis} showed how a passive unit and a sharing unit can relax the requirements of lightly geared actuators for a load-bearing lower-limb exoskeleton. A deeper analysis using off-the-shelf components will now show the actual benefits of the proposed design regarding the total actuation mass, inertia and efficiency. Suppose a strong multifunctional 2-DOF exoskeleton for infantry that can unload significantly its user's bodyweight (up to 100\% BW) when walking, allowing carrying heavy payloads. This exoskeleton could also assist (or at least not restrain) running, jumping and squatting when needed. The requirements are:
\begin{description}
    \item[\textbf{R1}] User Bodyweight: 75~kg
    \item[\textbf{R2}] Max 1.8~m/s Walking Assistance: 100\% BW
    \item[\textbf{R3}] Peak GRF (\textbf{R1}x\textbf{R2}): 1000~N/leg (fig.~\ref{fig:requirements_curves})
    \item[\textbf{R4}] Vertical Leg Stroke: 0.4~m
    \item[\textbf{R5}] Work-per-Stroke (\textbf{R3}x\textbf{R4}): 400~J/stroke/leg
    \item[\textbf{R6}] Max Jumping Speed: 2.8~m/s
\end{description}
The RMS force motor requirements of Table~\ref{table:requirements_RMS_peak} are used to design the dynamic force units. The entire transmission designs after the leader piston(s) are driven by \textbf{R5}. For speed, the jump task is the most restrictive, especially when sharing the same motor for both legs. We introduce the design variable $R_\text{motor-GRFy}$ that represents the total reduction ratio between the motor torque and the vertical force generated on the ground.

\begin{figure}[h!]
\centering
\includegraphics[width=1\columnwidth]{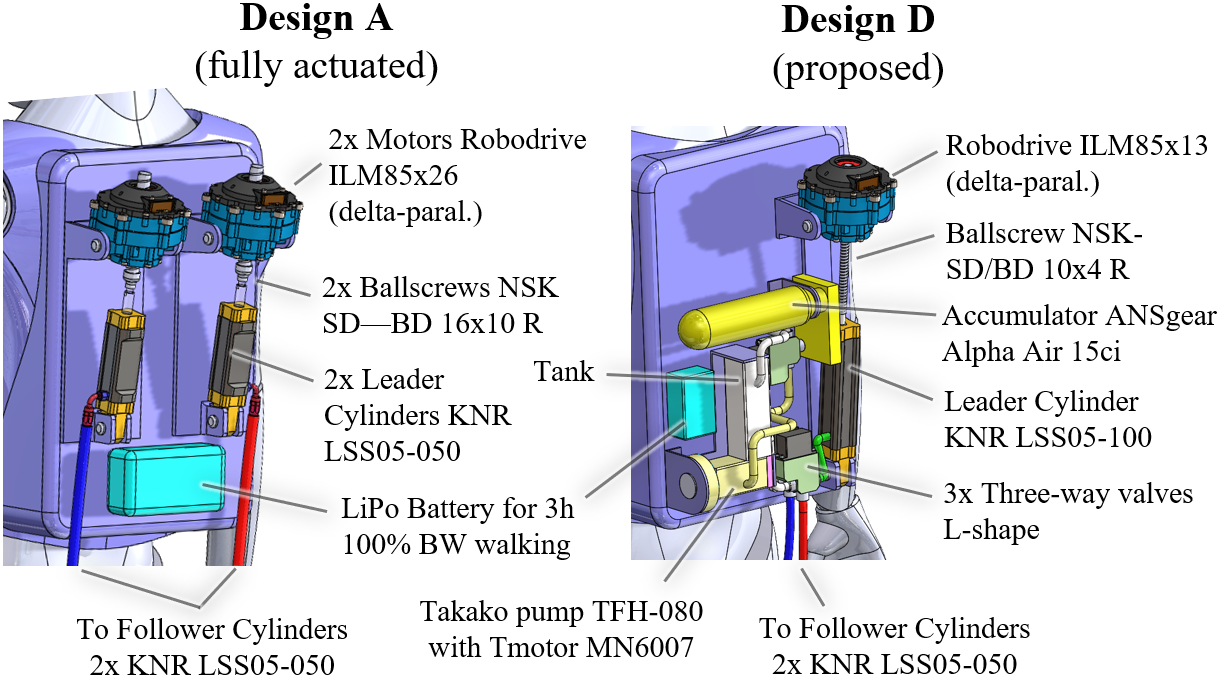}
\caption{Case study illustrative CAD. All actuation is located in the user back and actuates remotely the vertical GRF of the legs. See Table~\ref{table:caseStudyOuputs} for details regarding battery sizing.}
\label{fig:CaseStudyCAD}
\end{figure}

\begin{figure}[h!]
\centering
\includegraphics[width=1\columnwidth]{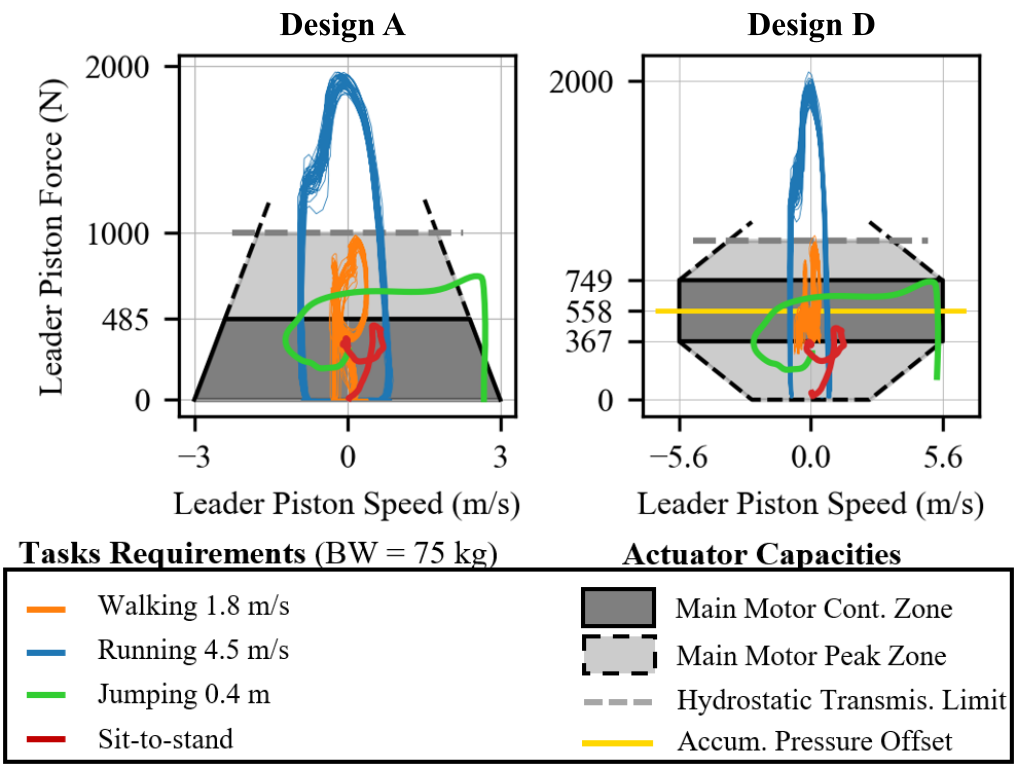}
\caption{Performance map coverage of the actuator designs from the case study, as seen in the vertical output reference frame. Both designs have enough speed to cover all tasks and each one can assist at a similar force amplitude a 75~kg BW user.}
\label{fig:CaseStudyPerformance}
\end{figure}

Fig.~\ref{fig:CaseStudyCAD} illustrates the resulting Designs A and D along with their performance maps in Fig.~\ref{fig:CaseStudyPerformance} to show how much force assistance each design can provide for each task.
For Design~A, the total ratio is $R_\text{motor-GRFy}=186~\si{\per\meter}$, whereas it is  $R_\text{motor-GRFy}=147~\si{\per\meter}$ for Design~D. With similar performance, the designs can be compared with Table~\ref{table:caseStudyOuputs} regarding mass, efficiency and inertia. Even with the extra components that Design D needs, the total mass is similar.

\begin{table}[h!]
\footnotesize
\vspace{5pt}
\begin{threeparttable}
    \centering
    \caption{Case Study Comparison}
    \label{table:caseStudyOuputs}
    \begin{tabular}{| l c c |}
  \hline
 \multirow[l]{2}{*}{\textbf{Performance}} & \textbf{Design A}& \textbf{Design D} \\
 & (fully actuated) & (proposed) \\
\hline \hline
 & & \\ [-1ex]
 \textbf{Power (W) and energy (J)}$^1$  & \textbf{77} & \textbf{20} \\ \hline 
 Main motor(s) heat power losses & 62 & 16 \\
 Ballscrew power losses & 15 & 4 \\
 Accumulator energy stored & - & 1570 \\
 
  & & \\ [-1ex]
 \textbf{Mass w/o Battery (kg)}  & \textbf{4.6} & \textbf{3.9} \\ \hline 
 Frameless Motor(s) & 1.34 & 0.40 \\ 
 Ballscrew(s) & 1.24 & 0.12 \\
 Leader Cylinder(s)$^2$ & 0.98 & 0.65 \\
 Follower Cylinders$^2$ & 0.98 & 0.98 \\
 Fluid & 0.07 & 0.31 \\
 Accumulator$^2$&   & 0.22 \\
 Motorized pump$^2$ &  & 0.61 \\
 Valves &  & 0.60 \\ \hline
 Battery mass per hour$^3$ (kg/h) & 0.50 & 0.07\\ 
  & & \\ [-1ex]
 \textbf{Reflected Inertia at each Leg (kg)} &  \textbf{3.8}  &  \textbf{1.4} \\ \hline
 Motor & 3.66 & 1.32 \\ 
 Ballscrew (nut rotating) &  0.15  &  0.10 \\  \hline
    \end{tabular}
     \begin{tablenotes}
      \footnotesize
      \item $^1$ At nominal motor(s) torque, i.e. the RMS powers for a 100\%BW continuous walking assistance at 1.8~m/s. Transmissions losses assuming 90\% ballscrew efficiency for both designs.
      \item $^2$ Selection of small-scale lightweight hydraulic components were used. Pumps and cylinders are commercially available for robotic applications. While lightweight composite accumulators have been commercialized in the past (SteelHead Composites MicroForce series), this case study uses a small carbon-fiber reservoir (20.7~MPa max working pressure, 6.9~MPa precharge pressure) due to the lack of currently available products.
      \item $^3$ Assuming 150~\si{\watt\hour\per\kg} LiPo batteries.
    \end{tablenotes}
    \end{threeparttable}
\end{table}
%

 
%

Regarding efficiency, Design D reduces power consumption by a factor of 3.9, which translates into increased autonomy or the use of a lighter battery pack. This 57~W power savings during use should be compared with the energy consumption of the pump. Assuming $\eta_\text{pump} \approx 75\%$, the pump would consume $\approx 2100$~J to pressurize the accumulator up to the pressure setpoint. This means the energy savings offset the initial charging cost after only 37~s of operation, making the passive system beneficial over relatively short usage periods. Design~D also consumes energy for valve switching, but it can be negligible\footnote{Only 0.13~J/switch for a custom-made design for an exoskeleton in \cite{fan_hyexo_2024}.}.

Regarding inertia, we note that, as a reference, the same motors are used for Design~A as in the backdrivable leg prosthesis of Elery et al. \cite{elery_design_2020} but here driving instead two ball screws and two hydrostatic transmissions to actuate each leg remotely. The reflected inertia of Design A is thus equivalent to this state-of-the-art backdrivable prosthesis designed for walking. For Design D, the resulting inertia is 2.7x lower, increasing the backdrivability, especially for powerful tasks like running. 

The actual mass/efficiency/inertia advantage of Design D over a fully actuated hydrostatic one varies with the requirements. Since it needs extra components, it will be more likely beneficial when a high force-to-inertia ratio is desired, e.g., for strong backdrivable robots. 

\section{Proof-of-Concept Design and Control}
\label{section:proof-of-concept}
This section presents the design of a proof-of-concept to validate the experimental feasibility of the proposed concept of Fig.~\ref{fig:proposed_concepts}. It also introduces the basic force control and switching control strategies for the validation.

\subsection{Test Bench Design}
\label{section:prototype_design}
\begin{figure*}[t]
\centering
\includegraphics[width=0.99\textwidth]{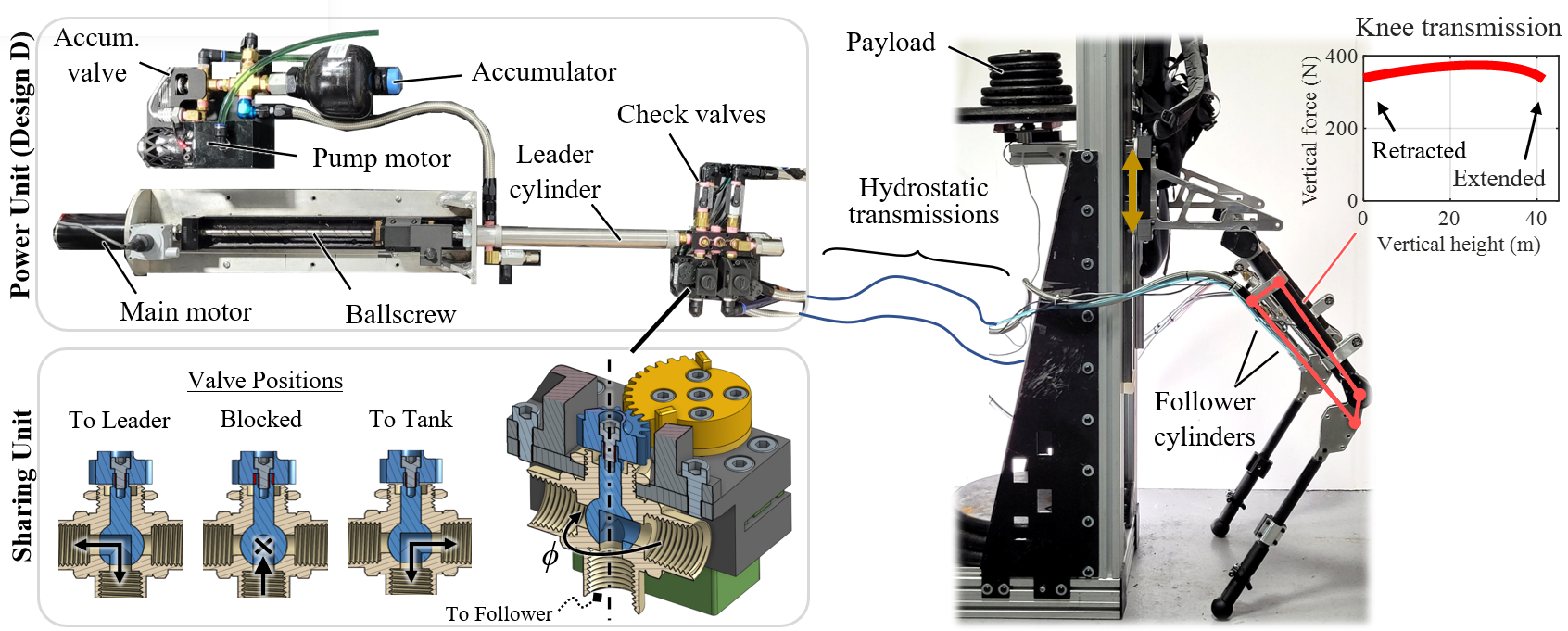}
\caption{Overview of the actuator proof-of-concept remotely powering the knees of an exoskeleton and its custom motorized ball valves.}
\label{fig:prototype_bench}
\end{figure*}

The proof-of-concept is a knee exoskeleton driven by a remote power unit on a table, as shown in Fig.~\ref{fig:prototype_bench}. Table~\ref{table:prototype} summarizes the specifications and the selected components.\\

\begin{table}[h!]
\footnotesize
\caption{Proof-of-concept components and specifications}
\label{table:prototype}
\centering
\begin{tabular}{| l c c |}
\hline
 & & \\ [-2ex]
\textbf{Description} & \textbf{Var.} & \textbf{Value} \\
 & & \\ [-2ex]
\hline
\hline
 & & \\ [-1ex]
\multicolumn{3}{| l |}{\textbf{System Specifications at Output} for 1 Leg [2 legs]} \\ 
\hline
Max Vertical Force (\si{\newton}) & - & 366 [711] \\
Max Dynamic force (\si{\newton}) & $F_\text{dyn,peak}$ & 81 [162] \\
Max Static Force (\si{\newton}) & $F_\text{stat.}$ & 366 [711] \\
No-load Speed (\si{\meter\per\second}) & - & 9.1 [4.6] \\
Reflected Inertia* (\si{\kg}) & - & 0.2 [0.4] \\
Single Leg Mass (\si{\kg}) & - & 3.0 \\
Legs + Guiding Rails Mass (\si{\kg}) & $m_\text{proto}$ & 13.4 \\
 & & \\ [-1ex]
\textbf{Main Motor} & & Maxon RE50 (48V)\\
\hline 
Torque constant (\si{\newton\meter\per\ampere}) & $K_\text{T}$ & 0.093\\
Torque (\si{\newton\meter}) & - & 0.42 cont.\\
 & & 1.4 peak \\
 & & \\ [-1ex]
 \textbf{Ballscrew (BS)} & & NSK MCM05025H20K00 \\
 \hline
 Lead  (\si{\milli\meter}/turn) & $L_\text{BS}$ & 20 \\
 Screw diameter (\si{\milli\meter}) & - & 12 \\
 & & \\ [-1ex]
\textbf{Cylinders} \scriptsize (L=leader, F=follower) & & Bimba H-093-DUZ \\
\hline
Max pressure (\si{\mega\pascal}) & - & 3.45 \\
Piston Area (\si{\milli\meter\squared}) & $A$ & 572 \\
Piston Area (rod side) (\si{\milli\meter\squared}) & $A_\text{r}$ & 524 \\
Strokes & $x_\text{L}$ & 203 \\
 & $x_\text{F}$ & 76 \\
 & & \\ [-1ex]
\textbf{Hydrostatic Transmission} & & SUM-230610 \\
 \hline
Internal Diameter (\si{\milli\meter}) & & 8.7 (-6 AN)\\
Length  (\si{\milli\meter}) & & 1000 \\
Fluid & & Recochem 35-365WP \\
 & & \\ [-1ex]
\textbf{Accumulator} & & Stauff STDA-0500\\
 \hline
Nominal (gas) volume (\si{\milli\liter}) & $V_{\text{a}0} $& 500 \\
Max pressure (\si{\mega\pascal}) & $P_\text{a}$ & 21 \\
Precharge pressure (\si{\mega\pascal}) & $P_{\text{a}0}$ & 1.03 \\
Max compression ratio & - & 1:8 \\
 & & \\ [-1ex]
\textbf{Gear Pump} & & 0AM325579D\\
\hline
Fluid & - & Pentosin CHF 11S \\
Pressure-Speed slope (\si{\pascal\second}) & $m_\text{p}$ & 9709 \\
  & & \\ [-1ex]
\textbf{Motorized Valves} & & McMaster 4149T42 \\
\hline
Max pressure (\si{\mega\pascal}) & - & 10.3 \\
Internal Diameter (\si{\milli\meter}) & - & 6.35 \\
Servomotor & - & AGFRC A80BHP-H \\
Stall Torque ($\si{\newton\meter}$) & - & 1.8 \\
0-180° switching speed ($\si{\milli\second}$) & - & 80 \\
0-113° switching speed ($\si{\milli\second}$) & - & 60 \\
 \hline
\end{tabular}
\begin{tablenotes}
\footnotesize
  \item *Including motor, screw and fluid transmission inertia.
\end{tablenotes}
\end{table}

\subsubsection{Exoskeleton and Transmission} The device is a two-link, non-anthropomorphic exoskeleton. The absence of ankle joints means the knees solely control the extension of the legs. The hip joints are free and the knee joints are actuated to assist the vertical GRF\footnote{We expect that the exoskeleton kinematics could be improved by decoupling completely the horizontal and vertical end-effector motions, so that the knee joint is dedicated solely to carrying the gravity loads, as in \cite{fan_load-carrying_2024}.}.

By deriving the kinematic equations of the robotic leg, the positions of the piston rod attachments (lever mechanism) were computed to minimize the variation of vertical force assistance with respect to leg vertical stroke at a given pressure, as shown by the plot on the upright corner of Fig.~\ref{fig:prototype_bench}. The prototype allows for different attachment configurations to change this output ratio on purpose. The configuration with maximum vertical stroke has an average ratio of the vertical rail force over the follower piston force of $\overline{R}=0.18$. The leader piston has twice the stroke of the follower pistons for having full leg stroke when both legs are connected to it. The transmission fluid was chosen for its low viscosity at room temperature, low toxicity and anti-corrosion properties. A low-viscosity fluid increases transmission efficiency and backdrivability, but this comes at the cost of an underdamped dynamic force response, as experimented in sections~\ref{section:experiments} and~\ref{section:switching_mitigations}.

\subsubsection{Adjustable Passive Force Unit}
\begin{figure}[h!]
\centering
\includegraphics[width=1\columnwidth]{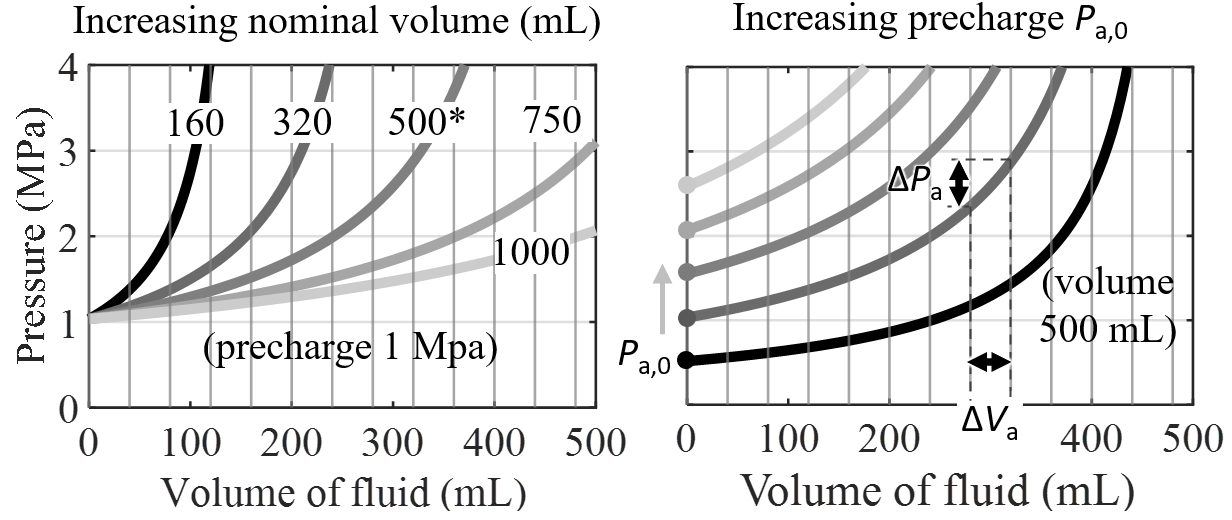}
\caption{Effects of the accumulator nominal (gas) volume and the gas precharge on the pressure-volume  curves. *Curve of the accumulator used for the experimental setup.}
\label{fig:accumulator_curves}
\end{figure}
Hydraulic accumulators have either a gas spring behavior (bladder and diaphragm accumulators) or a linear spring behavior (piston with a spring). A diaphragm accumulator is here selected for its low friction. For sizing the accumulator, two main factors influence its behavior: its nominal gas volume $V_{\text{a,}0}$ (size) and its gas precharge $P_{\text{a,}0}$. The relation between the accumulator pressure $P_\text{a}$ and volume $V_\text{a}$ is:
\begin{equation}
    P_\text{a}=P_{\text{a,}0} \frac{{V_{\text{a,}0}}^{n}}{\left( V_{\text{a,}0}-V_\text{a}  \right)^{n}}
    \label{eq:accumulator}
\end{equation}
where $n=1$ for isothermic change (slow) and $n=1.4$ for adiabatic (fast) change and for nitrogen gas. Fig.~\ref{fig:accumulator_curves} shows the effect of $V_{\text{a,}0}$ and $P_{\text{a,}0}$ on the pressure-volume relationship.
We want the quasi-static output force assistance not to vary much with leg stroke. When the leader piston moves, $V_\text{a}$ varies, inducing a variation of pressure $\Delta P_\text{a}$ as shown by Fig.~\ref{fig:accumulator_curves}. This effect can be reduced by:
1) using a high precharge $P_{\text{a,}0}$, but it decreases the available low-pressure range,
2) increasing the volume of the accumulator, but it increases its mass and the need for a powerful pump that fills it within a reasonable time,
3) designing the leg transmission to cancel out the effect of $\Delta P_\text{a}$ at output\footnote{This solution is analog to Fig.\ref{fig:gravity_compensation_concepts}a but is not working if the number of joints connected to the accumulator can change.}, or
4) increasing the overall system rated pressure because it decreases the displaced volume of the pistons (for the same work-per-stroke, less piston stroke or area is needed). The later point also means that the cylinders and accumulator are more compact for higher pressure rated systems. Here, a 500~mL gas volume and a 1~\si{\mega\pascal} precharge is chosen. The small car transmission gear pump can generate up to 3.5~\si{\mega\pascal} with the fluid suggested by the manufacturer. Its viscosity is higher than for the hydrostatic transmission but it has minor effects on backdrivability since this circuit is short.

\subsubsection{Dynamic Force Unit} This unit is a lightly geared actuator that generates 0.77~MPa in the transmission, at peak torque. The ball screw has a high lead which makes it backdrivable and the total ratio ensures having enough vertical speed for jumping with both legs connected. The hydrostatic transmission is filled using a high-flow diaphragm pump through multiple manual valves located along the circuits to simplify the filling procedure and minimize the trapped air.

\subsubsection{Sharing Unit}
A combination of two motorized three-way L-port ball valves enables to share the same actuator for the left and right knees. Solenoid valves would be faster but are heavy and restrict the flow. The leak-free design and high flow coefficient of ball valves make them suitable for reconfigurable hydrostatics but motorized industrial ball valves are too slow for most robotic devices. The prototype uses a high-speed servomotor designed for RC helicopter tail rotors. An additional 3D printed spur gear multiplier 1~:~2 ($R_\text{valve}=0.5$) increases the switching speed. Fig.~\ref{fig:prototype_bench} (left-bottom corner) shows the design. The mass of a valve unit is 247~\si{\gram} (valve 95~$\si{\gram}$, servomotor 79~$\si{\gram}$, plastic gears and frame and bolts 73~$\si{\gram}$). A full switch (180° stroke) takes $80$~ms but the output becomes partially connected within $60$~ms, at 113°. A more custom valve design 
 could improve the performance. For instance, for leg switching, \cite{fan_load-carrying_2024} designed recently a rotary-cage valve with a 24~ms rise time, consumes only 0.1~J/switch and weight 280 g but it is rated for low pressures.

\subsubsection{Software and Electronics Implementation}
The controllers were implemented on a Teensy 4.1 running at a consistent low-level frequency of 1000 Hz. Data acquisition occurs at 200 Hz on a Jetson Nano. The main motor is driven by a Maxon 70/10 with a current feedback loop, while the pump motor is controlled by an Odrive 3.6 using a speed feedback loop. Valve positions are monitored via 2048 PPR encoders (CUI AMT103). The robot's base height $y$ is measured with two ultrasonic range sensors (HC-SR04). Joint rotations are captured with 40000 PPR encoders (US Digital E6-10000-394-IE-D-D-D), and encoder signals are decoded through LS7366R quadrature counters. Pressure is monitored using Honeywell PX3AN1BS667PSAAX sensors. The leader piston position is tracked by a string potentiometer (TE Connectivity, SP2-12), and valve power consumption is measured with a TI INA260 power sensor.

\subsection{Controller}
\label{section:forceControl}
\begin{figure}[h!]
\centering
\includegraphics[width=0.49\textwidth]{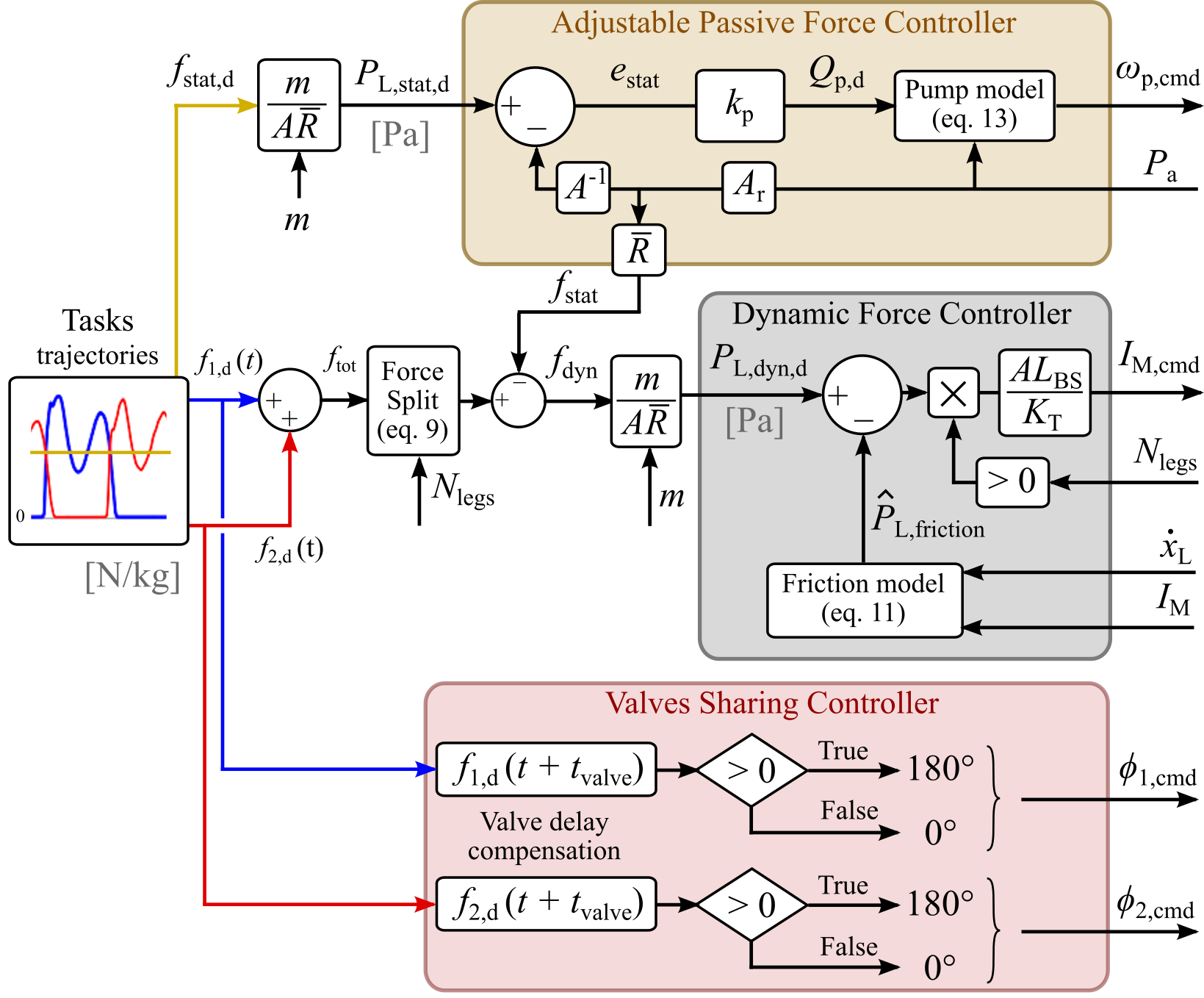}
\caption{Controller for sharing the reference forces between the main motor and the pump-accumulator unit and for connecting the legs to the actuator. Most parameters are defined in Table~\ref{table:prototype} or otherwise described in the text.}
\label{fig:force_controller}
\end{figure}

The controller presented in Fig.~\ref{fig:force_controller} is designed to track GRF trajectories of typical tasks while reducing the load on the electric actuators.
The philosophy behind the controller is to split force control into complementary static and dynamic components. The low-frequency forces are handled by the passive force unit controller (section~\ref{section:static_force_controller}). The high-frequency forces are handled by the dynamic force unit controller (section~\ref{section:dynamic_force_controller}). This separation allows the controller to minimize the load beared by the electric actuators. Finally, the sharing controller (section~\ref{section:sharing_controller}) coordinates the hydraulic valves to link the legs that need to generate ground reaction forces.

In practice, the desired leg forces ($f_{1\text{,d}}$ and $f_{2\text{,d}}$) and the static force ($f_\text{stat,d}$, precalculated for each task as described in Section~\ref{section:mass_efficiency_analysis} to minimize electric motor effort\footnote{A future implementation could compute the optimal $f_\text{stat,d}$ in real time to maximize efficiency over a given time window.}) are converted into a motor current command ($I_\text{M,cmd}$), a pump velocity command ($\omega_\text{p,cmd}$), and a valve angle commands ($\theta_\text{a,cmd}$, $\phi_\text{1,cmd}$, $\phi_\text{2,cmd}$). The forces are scaled according to the payload mass $m$ and the number of connected legs $N_\text{legs}$.


\subsubsection{Dynamic Force Unit Controller} An open-loop controller tracks the desired dynamic pressure $P_\text{dyn,d}$. In turns, the motor helps track sudden variations of the desired static pressure (e.g., payload $m$ varies rapidly) until $P_\text{L,stat,d}$ is reached by the pump. Finally, force tracking is improved by using a smoothed Coulomb-viscous friction model:
\label{section:dynamic_force_controller}
\begin{equation}
    \hat{P}_\text{friction} = \underbrace{  \Bigl( \mu_\text{seal} + (1-\eta_\text{BS})\left|\frac{I_\text{M}K_\text{T}}{L_\text{BS}A_\text{r}}\right| \Bigr) \text{tanh}\left(\gamma\dot{x}_\text{L}\right) }_{\text{Coulomb seal + ballscrew}} + \underbrace{b\dot{x}_\text{L}}_{\text{damping}}
\end{equation}

\begin{figure}[h!]
\centering
\includegraphics[width=1\columnwidth]{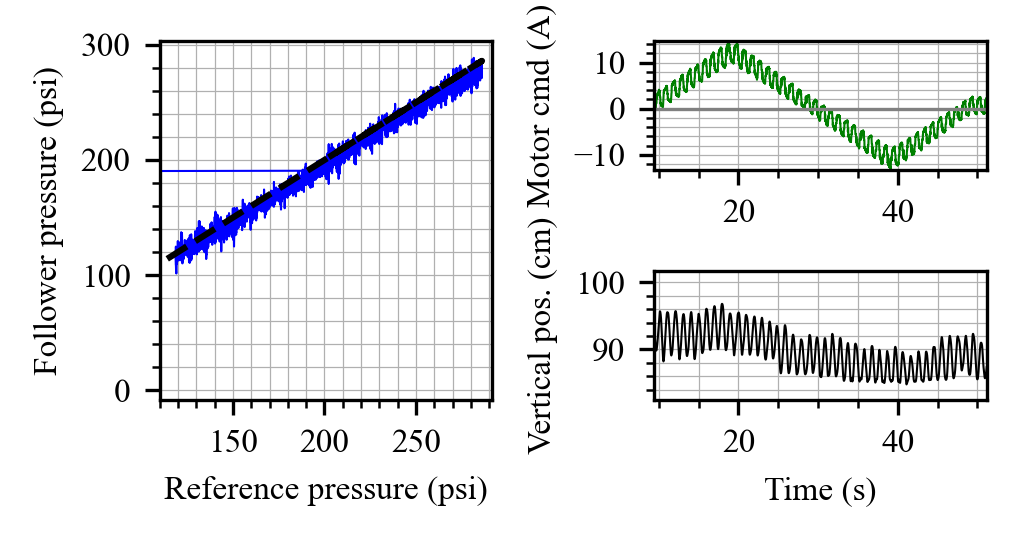}
\caption{Force tracking with friction compensation. The system is backdriven vertically at $\approx$1~Hz while an up-down reference force ramp is sent.}
\label{fig:force_control_with_without_compensation}
\end{figure}

For stable friction compensation, the friction parameters used are $\mu_\text{seal}=0.07$, $\eta_\text{BS}=0.96$ and $b=0.004$. The slope $\gamma$ is tuned experimentally to 0.3. %
Fig.~\ref{fig:force_control_with_without_compensation} shows the force tracking performance when the system is backdriven at 1~Hz.

\subsubsection{Passive Force Unit Controller} A pressure feedback controller compares the measured accumulator pressure $P_\text{a}$ to the desired leader static pressure $P_\text{L,stat,d}$. The gain $k_\text{p}=2.5$ converts this error into a desired gear pump flow $Q_\text{p,d}$. To find the required motor velocity command, the pump flow is given in a simplified form as \cite{logan_t_williams_fundamentals_2022}:
\label{section:static_force_controller}
\begin{equation}
    Q_\text{p}=V_\text{displ.}\omega_\text{p}-(\underbrace{\alpha\Delta P_\text{p}-\beta\omega_\text{p}}_{ Q_\text{leak}})
    \label{eq:model_pump}
\end{equation}

$V_\text{displ.}$ being the volumetric displacement.
The axial and radial leak flows through the gears $Q_\text{leak}$ increase with the pump pressure $\Delta P_\text{p}$ and decreases with the pump velocity $\omega_\text{p}$. $\alpha$ and $\beta$ depend on the geometry and the viscosity of the fluid. From equation~\ref{eq:model_pump}, the pump control law becomes:

\begin{equation}
    \omega_\text{p,cmd} = \underbrace{ \left( \frac{1}{V_\text{displ.}-\beta} \right)}_{m_Q} Q_\text{p,d}  + \underbrace{ \left(  \frac{\alpha}{V_\text{displ.} - \beta} \right)}_{1/m_P} P_\text{a}
    \label{eq:model_pump_law}
\end{equation}

where the coefficients $m_\text{P}\approx9709~\si{\pascal\second}$ and $m_\text{Q}\approx9.5~\si{\per\second\per\milli\liter}$ were found experimentally.
%
When the static pressure is reached, the accumulator circuit is blocked with a valve and the pump stops.

\subsubsection{Sharing Unit Controller}
\label{section:sharing_controller}
Leg switching is achieved by commanding the valve servomotors to move between two reference positions: the tank position (0°) when the desired force is zero, and the leader piston position (180°) when both desired forces are positive. These reference commands are tracked by the embedded servomotor position controllers. Additionally, the valves can be set to intermediate states (see Fig.~\ref{fig:prototype_bench}) to dissipate energy and smooth transitions as tested in section~\ref{section:transientsValves}. The valves takes $t_\text{valve}=60~\text{ms}$ to change position. The target force trajectories being pre-programmed in the experimental validation of this work, the valve commands are triggered in advance (see Fig.~\ref{fig:force_controller}). This ensures that the actuator is synchronized with the leg at the correct time. Future human trials would require online gait estimation, passive triggering, or faster valves, as discussed later.

In the next section, the prototype is operated to assess the performance of the proposed concept in various task scenarios.

%
%


\section{Experimental Validation}
\label{section:experiments}
This section presents preliminary experimental tests suggesting that the proposed actuator can meet the expected performance of a variety of vertical GRF profiles tasks for a load-bearing robotic leg that could eventually be used as an exoskeletons. Some undesired effects are found and discussed, such as relative to leg switching. Also, an energy consumption test is conducted and suggests that the proposed actuator can consume up to 4.8x times less energy for walking. The robot is guided along a vertical rail. The demonstrations were filmed and presented in the video attached to the paper. Table~\ref{table:validationRefernces} contains all the parameters and references used for the tests.

\begin{table}[h!]
\footnotesize
\renewcommand{\arraystretch}{1.1}
\caption{Parameters and References Used for the Validation Tests}
\label{table:validationRefernces}
\centering
\begin{tabular}{| c  c | c | c | c | c | c | }
\multicolumn{7}{c}{} \\ [-2ex]
\multicolumn{1}{c}{} & \multicolumn{1}{c}{} & \multicolumn{1}{c}{\textbf{Knee}} & \multicolumn{1}{c}{\textbf{Beared}} & \multicolumn{1}{c}{\textbf{Right}} & \multicolumn{1}{c}{\textbf{Left}} & \multicolumn{1}{c}{\textbf{Static}} \\

\multicolumn{1}{c}{} & \multicolumn{1}{c}{} & \multicolumn{1}{c}{\textbf{ratio}} & \multicolumn{1}{c}{\textbf{Mass}} & \multicolumn{1}{c}{\textbf{leg}} & \multicolumn{1}{c}{\textbf{leg}} & \multicolumn{1}{c}{\textbf{offset}} \\

\hline
\multicolumn{2}{|c|}{\multirow{2}{*}{Task}} & \multirow{2}{*}{$\overline{R}$} & $m$ & $f_\text{1,d}$ & $f_\text{2,d}$ & $f_\text{stat,d}$ \\ 
 & & & \scriptsize[\si{\kilo\gram}] & \scriptsize[\si{\newton\per\kg}] & \scriptsize[\si{\newton\per\kg}] & \scriptsize[\si{\newton\per\kg}] \\
\hline
\hline 
\multicolumn{2}{|c|}{Squatting} & \multirow{2}{*}{0.18} & $m_\text{proto}+$ & \multirow{2}{*}{$g/2$} & \multirow{2}{*}{$g/2$}  & \multirow{2}{*}{$g/2$}  \\
 & & & $m_\text{load}$ & & & \\
\hline 
\multirow{4}{*}{\rotatebox[origin=c]{90}{Jumping}} & charge & \multirow{4}{*}{0.18} & \multirow{4}{*}{$m_\text{proto}$} & $0$ & $0$ & \multirow{4}{*}{24} \\
& launch & & &  24 & 24 & \\
& in air & & & $0$ & $0$ & \\
& land & & & 24 & 24 & \\
\hline
\multicolumn{2}{|c|}{Walking} & 0.18 & $m_\text{proto}$ & Fig.~\ref{fig:requirements_curves} & Fig.~\ref{fig:requirements_curves} & 8.2$^{**}$ \\
\hline
\multicolumn{2}{|c|}{Running} & \multirow{2}{*}{0.18} & \multirow{2}{*}{0} & Fig.~\ref{fig:requirements_curves} & Fig.~\ref{fig:requirements_curves} & \multirow{2}{*}{8.5$^{**}$}\\
\multicolumn{2}{|c|}{\scriptsize(60\% assist.)} & &  & (60\%) & (60\%) &  \\
\hline
\multicolumn{2}{|c|}{Walking} &\multirow{2}{*}{0.34} & \multirow{2}{*}{$m_\text{proto}$} & \multirow{2}{*}{Fig.~\ref{fig:requirements_curves}} & \multirow{2}{*}{Fig.~\ref{fig:requirements_curves}} & \multirow{2}{*}{ 8.2$^{**}$} \\
\multicolumn{2}{|c|}{\footnotesize(energy test)} & & &  &  & \\
\hline
\multicolumn{7}{l}{\footnotesize $^*$Leading to a max pressure reference in the transmission (3.45~\si{\mega\pascal}).} \\
\multicolumn{7}{l}{\footnotesize $^{**}$Optimal values computed in section~\ref{section:mass_efficiency_analysis}} \\
\end{tabular}
\end{table}
%
%
%
%
%

\subsection{In-Phase Tasks Force Tracking (Combined Legs)}
In-phase tasks are tasks where both legs work together and have a similar force profile. Here are validated squatting (or similarly sit-to-stands) and jumping.

\subsubsection{Squatting and Adjusting to Varying Payloads} 
\begin{figure}[h!]
\centering
\includegraphics[width=0.99\columnwidth]{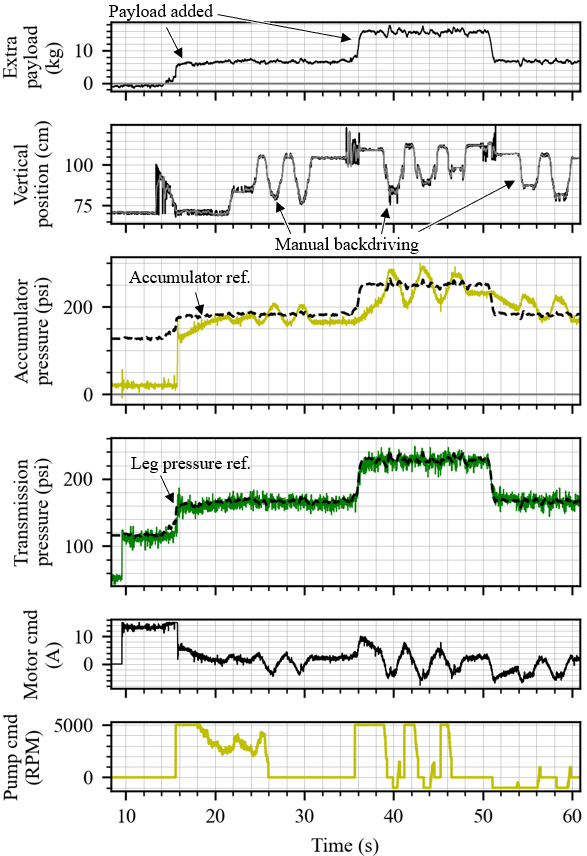}
\caption{Squats with varying payload validation. Pressure references in black, pump-accumulator signals in yellow, leader cylinder and motor signals in green, right leg signal in blue and left leg signal in red.}
\label{fig:squat_results}
\end{figure}

This test validates that the system and controller can balance gravity force and even automatically adapts to a varying payload ($m_\text{load}$) plus the weight of the vertically guided prototype. The exoskeleton is also manually backdriven up and down to simulate a squat-like motion. Fig.~\ref{fig:squat_results} shows that the system can automatically pressurize the accumulator to passively balance a measured load.
In parallel, the dynamic force controller meets three functions: 1) tracking instant variations of the payload until the accumulator is filled (at 16~s, 36~s and 50~s), 2) canceling out pressure variations $\Delta P_\text{a}$ due to the volume variations in the accumulator $\Delta V_\text{a}$ induced by the leader piston being backdriven back-and-forth (at 24~s, 38~s and 54~s), and 3) compensating friction.

Squat-like motions were also performed with higher payloads\footnote{Payload up to 43.2~kg plus the 13.4~kg mass of the exoskeleton, performing squat-like motions using manual control inputs only.} (see supplementary video), but the electric motor was not strong enough for proper compensation of friction and accumulator pressure variation.
%
\subsubsection{High Power Jumping} Most high force exoskeletons hinder powerful dynamic movements such as jumping because they lack mechanical transparency, have limited actuator speed and have too heavy collocated weight along the legs. The proposed actuator can generate high power using the passive force unit. The jumping sequence is: 1) charge the accumulator while being crouched with the leader piston fully retracted, 2) acceleration until the legs are fully retracted, i.e. when the height of the robot base $y<110$~cm, 3) aerial phase when $y>110$~cm where the exoskeleton must not constraint the user's natural movements, 4) land, again when $y<110$~cm.
\begin{figure}[h!]
\centering
\includegraphics[width=0.99\columnwidth]{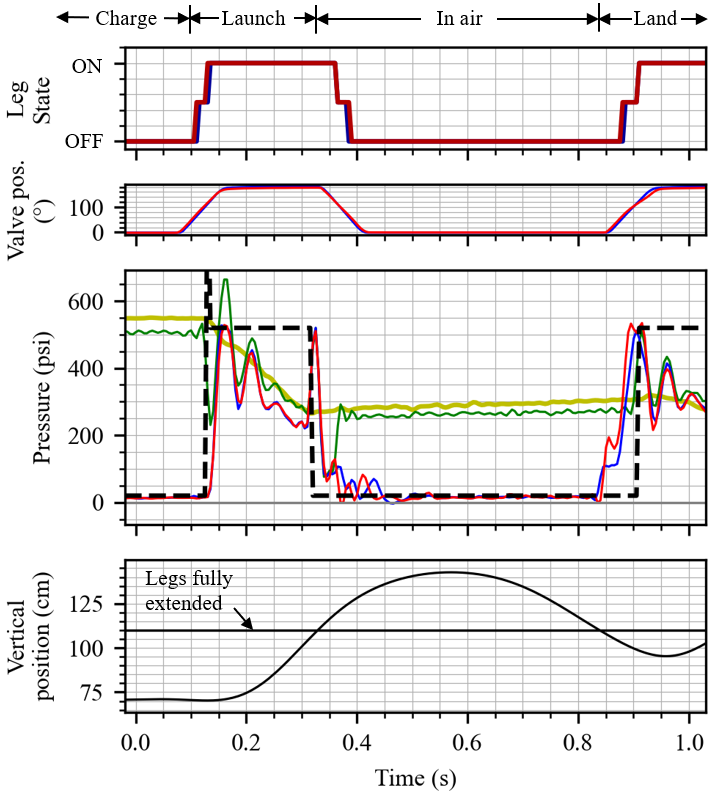}
\caption{Jumping validation. Pressure reference in dashed black, accumulator in yellow, leader cylinder in green, right leg in blue and left leg in red.}
\label{fig:jump_results}
\end{figure}
Fig.~\ref{fig:jump_results}
shows a 33~cm jumping height reached, which is near the mean human jumping height. The maximum vertical speed is 3.15~m/s and is near the maximal speed of the motor. During launching, the maximum pressure is not maintained because the dynamic force unit, even at peak torque, cannot overcome the accumulator pressure drop. Then, the valves are fast enough to disconnect the legs from the dynamic and static force units for 88\% of the aerial phase. Regarding power, we find during the launching sequence that the average and peak total leg powers are 790~W and 1220~W, respectively (using the slave pressure and the vertical speed signals), while the average and peak powers from the main motor are only 337~W and 507~W (using the motor current and the leader piston speed). These results suggest that the proposed topology extends the force and power capabilities of a low-force lightly geared motor, potentially allowing a user to generate high power and jumps naturally without being hindered. This full power assistance requires a delay to fill the accumulator, though. Note also that the supplementary video depicts post-landing bouncing, as the controller simply targets a high force. A more refined sequence would dampen the landing impact by modulating the force -- either using the motor or the valves to dissipate energy.

\subsection{Out-of-Phase Tasks Force Tracking (Alternating Legs)}
Out-of-phase tasks are when the legs support the load alternatively like for walking and running. These are more challenging because the hydrostatic topology is actively reconfigured which is discussed in this section.

\subsubsection{Walking and Running Force Tracking}
For walking at 1.8~\si{\meter\per\second} and running at 4.5~\si{\meter\per\second}, the references are the ground reaction force profiles of the first column of Fig.~\ref{fig:requirements_curves} since it corresponds to the output force to generate by each leg. For walking, the whole weight of the exoskeleton is beared by the leg in the stance phase. For running, it is in blocked position since the prototyped actuator cannot generate enough force to jump on one leg. It provides here a 60\% running force assistance. For both validations, the swinging leg is hydraulically disconnected from the actuator. However, since the user did not wear the exoskeleton in these preliminary demonstrations, no swinging motion is done.

\begin{figure}[h!]
\centering
\includegraphics[width=0.99\columnwidth]{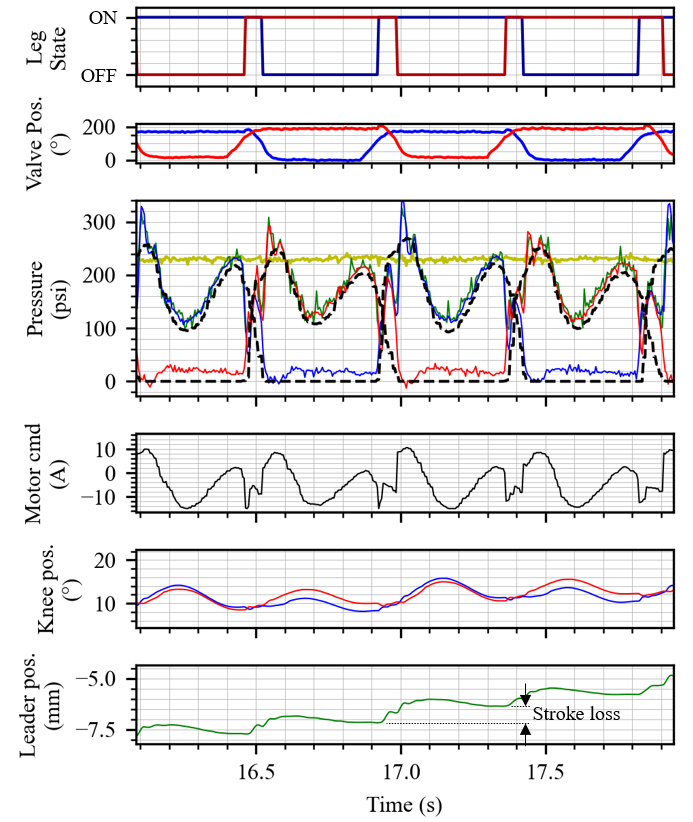}
\caption{Walking validation (1.8 m/s force profile) with the exoskeleton self-standing and free to move. Pressure reference in dashed black, accumulator in yellow, leader cylinder in green, right leg in blue and left leg in red.}
\label{fig:walk_results}
\end{figure}
\begin{figure}[h!]
\centering
\includegraphics[width=0.99\columnwidth]{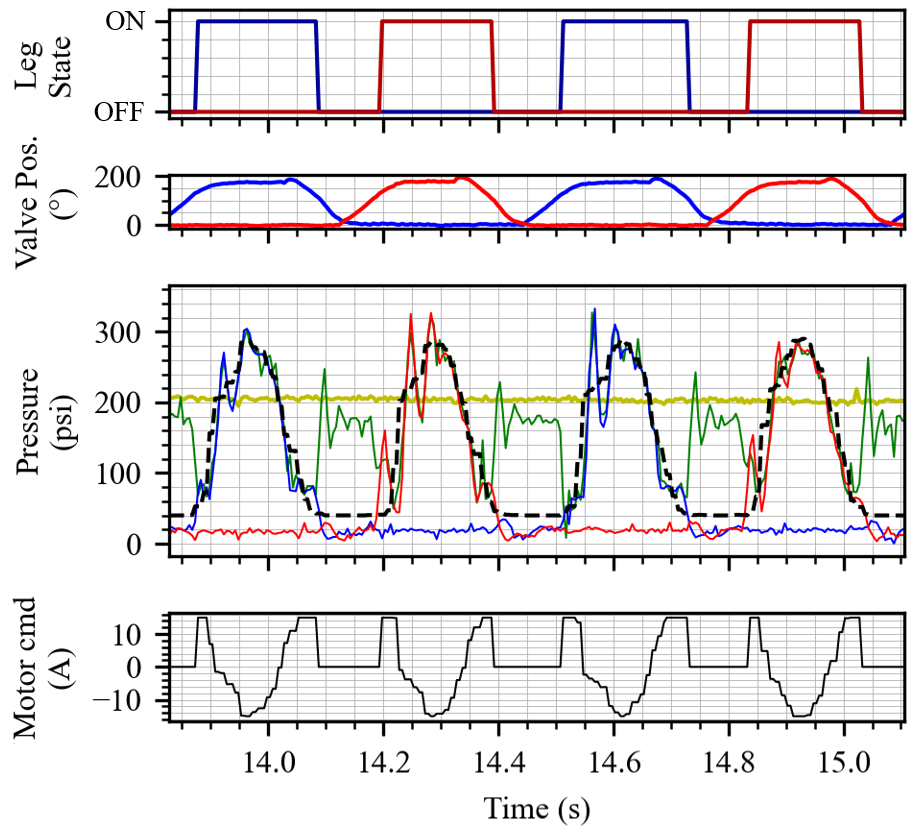}
\caption{Running force tracking validation (4.5~m/s force profile) with the exoskeleton legs blocked. Pressure reference in dashed black, accumulator in yellow, leader cylinder in green, right leg in blue and left leg in red.}
\label{fig:run_results}
\end{figure}
The results are given in figures~\ref{fig:walk_results} and \ref{fig:run_results}. For walking, the resulting exoskeleton motion is mostly a sine wave, like the actual motion of the center of mass of a human walking. The tests show that the right and left force profiles can be tracked alternatively, even for fast dynamic tasks like running. They also highlight a few limitations and challenges that are discussed next.
\begin{figure*}[t]
\centering
\includegraphics[width=0.99\textwidth]{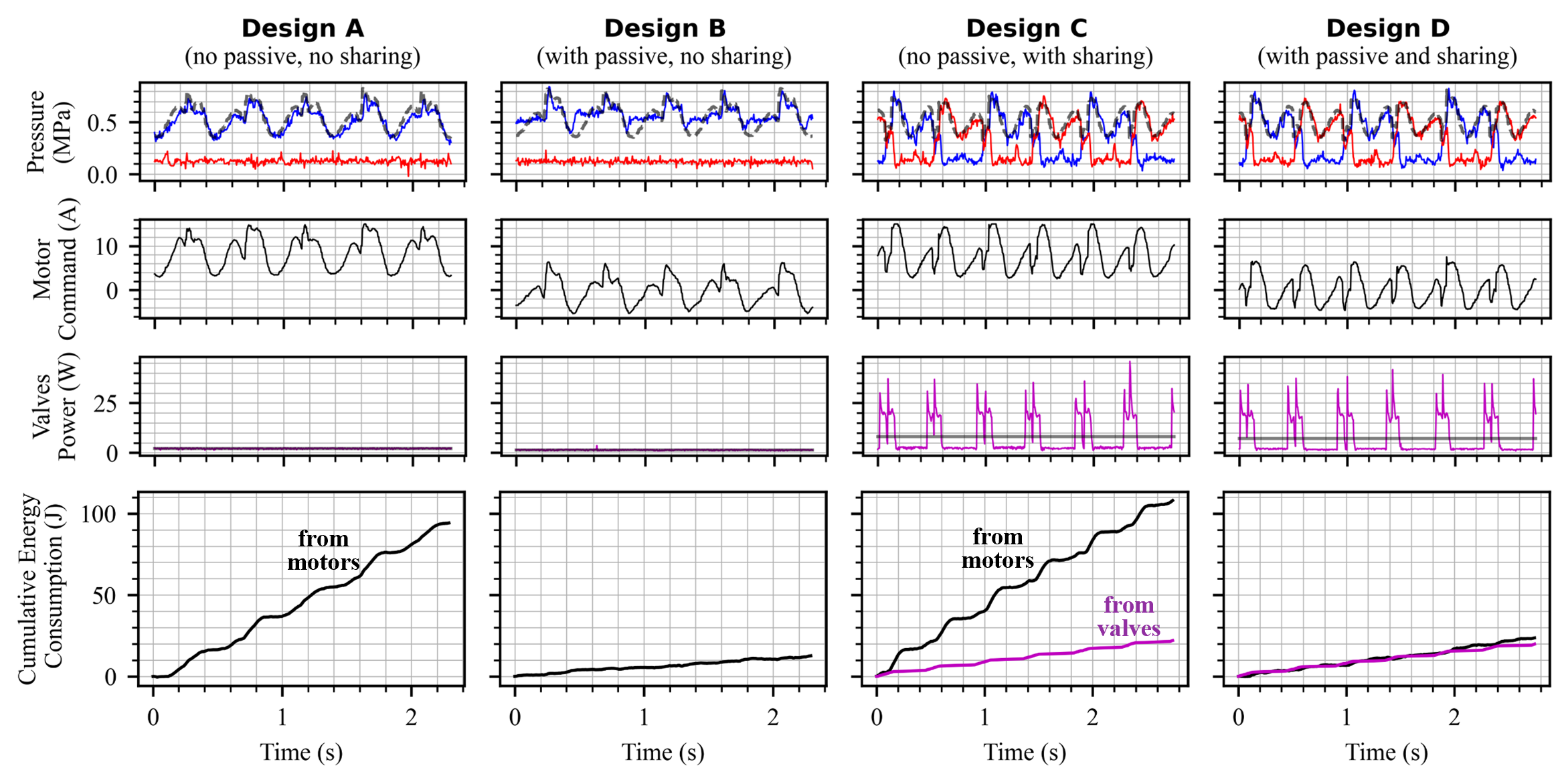}
\caption{Energy consumption tests for Designs~A--D when tracking the vertical GRF of walking. To imitate Designs~A and B (one motor per leg), the single motor tracks the total GRF of both legs with the right leg only.}
\label{fig:energy_results}
\end{figure*}
\subsubsection{Stroke Loss}
For walking, as annotated in Fig.~\ref{fig:walk_results}, some leader piston stroke is lost every step (here, 0.75~mm/step). This potential elastic energy is lost in the tank when a leg still under pressure gets disconnected which happens during the double support phase of walking. To avoid this, we shall not connect both legs simultaneously during walking, e.g., by prioritizing the assistance of the leading leg only, just like existing walking underactuated exoskeletons do \cite{asbeck_multi-joint_2015,fan_load-carrying_2024}. Surprisingly, stroke losses also happened for the running test but not as much (0.20~mm/step). This is for the same reason: the hydrostatic transmissions were still slightly pressurized when legs got disconnected, probably here due to wrong valve timing. Proper valve timing is thus essential. A pressure condition could be added to the valve control law to ensure the pressure is released before switching. Another mitigation to prevent missing stroke could be 1) replacing the leader piston by a hydraulic actuator with unlimited stroke, like a novel backdrivable and efficient pump, or 2) modify the actuator topology so that the pump can refill the hydrostatic transmission when needed.


\subsubsection{High Frequency Pressure Oscillations}
For running and walking, oscillations that match the natural transmission frequency occur when the legs are connected and when the main motor commands vary instantly, that is, when $N_\text{legs}$ changes discretely between 1 and 2 in equation~\ref{eq:C_motorForce}. It is not clear if these oscillations would be perceptible to a user, though. To prevent this, mechanical damping characteristics could be obtained by designing a valve with special-shaped orifices. Otherwise, model-based control laws for hydrostatic transmissions may be used to minimize overshoots and keep a high control bandwidth \cite{denis_low-level_2021}. The oscillations can also be mitigated by actively dissipating energy with the valves which is explored in section~\ref{section:transientsValves}.

\subsubsection{Valve Switching Delay}
For walking and running, the valves were switched $\approx$~65~ms in advance so that the legs get connected at the right time. For a realistic use-case scenario, we would need to estimate the percentage of the gait cycle in real time to synchronize leg connections. Moreover, the tests do not show whether the swinging leg would be hampered by the valve momentarily blocking the transmission during switching. In fact, if the valves are too slow or the transmissions very stiff, the user may feel resistance if he compresses the transmission during switching. This is addressed in section~\ref{section:transientsValves}. However, if the user pulls on the transmission while switching, he would not feel resistance because hydrostatic transmissions cannot transmit tensile forces.

Despite these challenges, vertical GRF tracking for walking and running appears feasible. Next, we compare the experimental energy consumption of the proposed concept with that of a fully actuated one to verify the efficiency advantage.

\subsection{Energy Consumption Comparison for Walking}
This test compares the experimental energy consumption of Designs A--D to assist in the stance phase of walking at 1.8~m/s. Therefore, the theoretical efficiency benefit calculated in section~\ref{section:mass_efficiency_analysis} is verified even in the presence of additional sources of energy loss and under at least one specific testing condition.

The main motor of the prototype is not capable of lifting the whole prototype on one leg without the passive force unit, which is necessary to simulate Designs~A and C. Then, the knee ratio is increased here to $\overline{R}=0.34$ by changing the configuration of the knee transmission (moving the attachment point of the follower cylinders). For the same reason, the accumulator precharge pressure is changed to 0.48~\si{\mega\pascal}. For testing Designs A and C (no passive force unit), the accumulator is disconnected using its dedicated valve. To reproduce Designs A and B, all the GRF references ($f_\text{1,d}+f_\text{2,d}$) are sent to the right leg only because the prototype just have a single motor.

The results are given in Fig.~\ref{fig:energy_results}.
The average power consumption from the main motor for Designs A--D are 41~\si{\watt}, 5.5~\si{\watt}, 39~\si{\watt} and 8.6~\si{\watt}, respectively. Compared to Design A (baseline), the motorization for Design B consumes 7.5x less, similar for Design~C and 4.8x less for Design~D. 
By comparison, the hydrostatic static force unit in \cite{fan_load-carrying_2024} reduced the energy consumption of walking by 65\% (including swing phasein this case).
%
%
The average power consumed by the switching valves is $\approx$~7~\si{\watt}, or $\approx$~3.5~\si{\joule} per full leg switch, which is modest compared to Design~A’s motor consumption. Still, custom low-power valves, as in \cite{fan_load-carrying_2024}, could further reduce this overhead and improve switching speed. As for accumulator pressurization, the pump consumed only 110~J (at the power supply) in this case, since the setpoint was close to the precharge pressure\footnote{For reference, a full pressurization from 0.9 to 3.8~MPa, as for the jumping validation, required 2100~J at the power supply.}. Thus, motor power savings become significant quickly. 

In conclusion, it is reasonable to think that the proposed actuation technology is feasible, multifunctional and more efficient. Careful attention is needed for underactuation during the double support phase of walking to prevent missing stroke. Pressure oscillations and momentarily blocking valves issues are addressed in section~\ref{section:transientsValves}. \par
\section{Challenges Mitigation Using the Valves}
\label{section:switching_mitigations}

%

\label{section:transientsValves} Many exoskeleton assisting devices that feature discrete transmission modes (e.g., two-speed gear transmissions, clutchable springs, etc.) introduce delays and parasitic effects that can be felt by the user when switching. As observed in Section~\ref{section:experiments}, switching can cause parasitic pressure transients. This section provides further insight into these effects, showing that both valve switching speed and output motion during the transition influence pressure overshoot. A few improvement strategies are evaluated, but only under simplified test conditions that isolate the problematic situations, rather than within the main experiments of the previous section. The results are promising and warrant further investigation in future work.

\subsubsection{Valve Switching Speed}
\begin{figure}[h!]
\centering
\vspace{-15pt}
\subfloat[\centering]{{\includegraphics[height=2.1in]{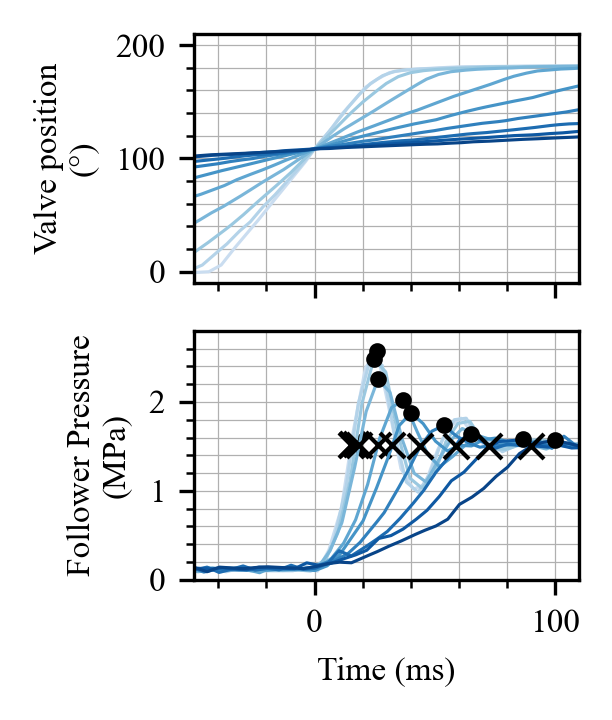}}}%
\qquad
\subfloat[\centering]{{\includegraphics[height=2.1in]{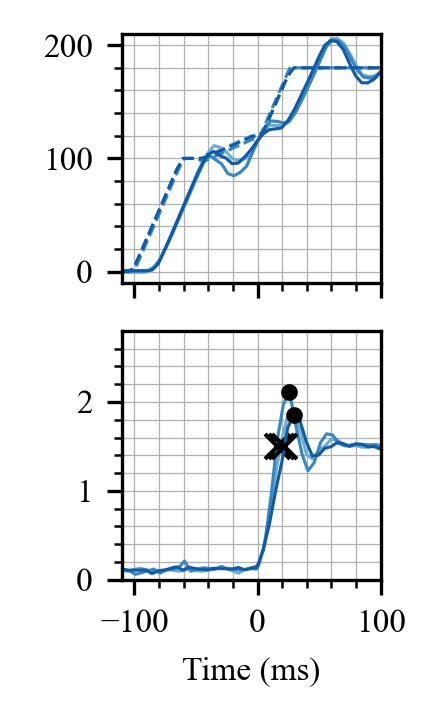}}}%
\caption{Effect of the valve speed when connecting the right leg (blocked output) to the leader cylinder under pressure: a) at fixed switching speeds, b) at variable switching speeds. All tests were synchronized in post-processing where the pressure begins to rise. The dots are the peak values and the crosses give the 0-100\% rising time.}
\label{fig:variableSpeedswitching}
\end{figure}

When the hydrostatic transmission is suddenly connected to a pressurized actuator, the pressure inside the line overshoots before stabilizing. Pressure feedback could hardly be used to dampen these oscillations with the main motor since it matches the natural frequency of the transmission. Here the effect of valve switching speed is assessed for the case of a static output. The right leg is blocked and the leader cylinder is pressurized at $1.5~\si{\mega\pascal}$ by the accumulator. The right valve is then switched. This procedure is repeated for ten different valve speed values ranging from $2600\si{\degree\per\second}$ (max speed) to $180\si{\degree\per\second}$. The results in Fig.~\ref{fig:variableSpeedswitching}a show that switching slower can help dampen the second order dynamics of the transmission, but at the compromise of longer switching times. 
%
This is due to the partially blocked position of the valve (see Fig.~\ref{fig:prototype_bench}) where the head loss through the valve is high. Most hydraulic valves have a similar middle position with high head loss.

The motorized valve can be seen as an extra force input to the dynamic system to dissipate energy on purpose. We propose to vary the valve speed during switching, as demonstrated in Fig.~\ref{fig:variableSpeedswitching}b. The valve speed is maximal for most times but is greatly reduced when the valve position is near its partially blocked state. The compromise between switching speed and pressure overshoot is then reduced.

\subsubsection{Switching with a Moving Output}
\begin{figure}[h!]
\centering
\vspace{-15pt}
\subfloat[\centering]{{\includegraphics[height=2.0in]{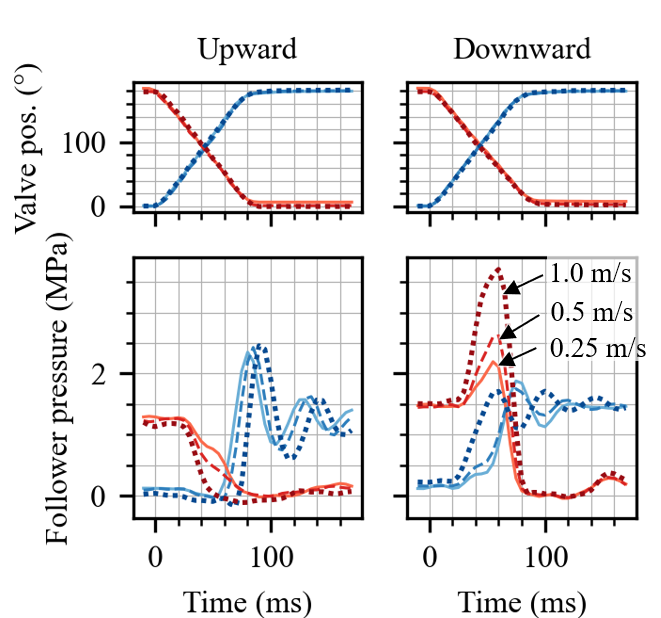}}}%
\qquad
\subfloat[\centering]{{\includegraphics[height=2.0in]{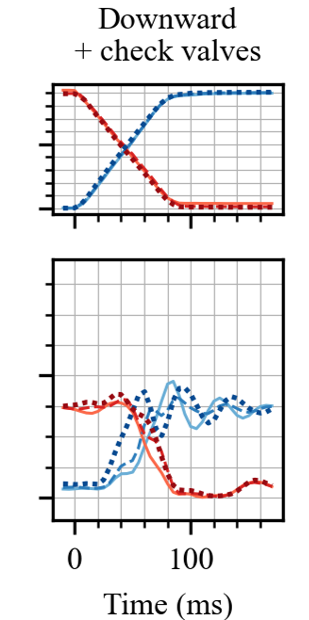}}}%
\caption{Leg switching tests at different output cart velocities (0.25~m/s = solid lines), 0.50~m/s = dashed lines and 1.00~m/s = dotted lines): a) when the output is moving upward and downward, b) solution for improving the downward switch using check valves to by-pass the flow. Blue and red lines are for the right and the left leg measurements, respectively.}
\label{fig:switchingTests_variableVerticalSpeed}
\end{figure}

When switching, the valves are blocked for a few milliseconds (see Fig.~\ref{fig:prototype_bench}). Meanwhile, the leg becomes passive and the output force is driven by the motion of the output and the stiffness of the hydrostatic transmission. This could hinder the user at the transitions of tasks as when landing a jump or between the swing and stance phases of walking/running. The tests of Fig.~\ref{fig:switchingTests_variableVerticalSpeed} illustrate this effect. The weight of the vertical cart is first balanced by the accumulator. Then, the cart is manually moved in one direction and leg switching is triggered when a given vertical speed condition is met (upward and downward motions at 0.25~m/s, 0.5~m/s and 1.0~m/s). The valves switch at maximum speed. A series of tests were also conducted for a slower switching speed (200~ms) but are not presented here for conciseness.
On one hand, switching while the output is moving upward has minor effects on the user. Indeed, the line depressurizes in advance, down to $\approx$~0~MPa (vacuum), removing the force applied to the output/user but not restricting its motion. On the other hand, when switching during a downward motion, the valve restricts the motion of the output/user. Actually, the faster the output is moving and the slower the valves  are (and the stiffer the transmission is), the higher the pressure overshoot is. To mitigate this effect, we propose to add two optional check valves in parallel to the leg valves (see Fig.~\ref{fig:prototype_bench}) to by-pass them when they are in the blocked position. Check valves are typically small and inexpensive components. As expected and shown by Fig.~\ref{fig:switchingTests_variableVerticalSpeed}b, this removes the overshoot when the output is moving downward. A beneficial side effect is also that the leg gets connected faster through the check valves, independently of the valve switching speed performance. 

\section{Discussion and Conclusion}
Fully actuated robots using lightly-geared motors are multifunctional but generate heat and require heavy motorization. Quasi-passive and underactuated robots are more efficient and use fewer motors, respectively, but are typically task-specific. This paper presents a reconfigurable hydrostatic concept combining adjustable static load compensation with a sharing mechanism via hydrostatic transmissions. This hybrid concept is multifunctional. In exoskeletons, the sharing mechanism is novel in that it can assist both in-phase movements (squats and jumps), and out-of-phase movements (walking, running). The adjustable static load compensation can be applied to load-bearing robotics in general, providing real-time adaptability for varying payloads or bodyweight support.

\textbf{Theoretical Benefits:} The analytical studies highlighted potential reductions in motorization requirements and improvements in efficiency, mainly due to the static load compensation. A case study showed that, despite added complexity, actuation weight remains similar while increasing backdrivability and efficiency. The proposed concept is expected to stand out for stronger, more transparent, and long-autonomy load-bearing robotics, as saved motor and battery mass can offset the mass overhead of hydraulic components. Lighter miniature hydraulic components would further enhance performance.

\textbf{Experimental Validations:} The multifunctionality was validated by tracking the vertical GRF of various tasks with a proof-of-concept actuator driving a self-standing prototype. Stance phases of both knees could be actuated with a single actuator through the hydrostatic sharing mechanism. Further tests reproducing walking GRF indicated a 4.8-fold reduction of energy consumption under specific conditions. While not yet generalizable, these results indicate that theoretical advantages can be realized despite additional losses from valves and hydraulics. Finally, low-viscosity fluid for better transmission efficiency and backdrivability led to force oscillations upon leg switching. Active energy dissipation using the valves is promising to mitigate these underdamped force responses. Check valves are also promising to passively connect the legs upon ground contact.

\textbf{Method Limitations:} The proposed design performance was assessed with a proof-of-concept actuator driving a self-standing robotic leg based on pre-computed force trajectories only. Hence, key human-robot interaction effects — comfort, gait adaptation, skin contact pressures — were not captured. Adaptative online valve synchronization and static force optimization were also not addressed.

\textbf{Future work:} Human trials are needed to ensure comfortable switching and effective assistance. The explored valve mitigation should be further tested as a mean to reduce switching effects and to enable passive leg connection. Concept simplifications inspired by recent works \cite{shveda_wearable_2022, fan_hyexo_2024} could eliminate the hydraulic pump, using the cylinder as a pump to harvest locomotion energy to adjust the accumulator’s assistive pressure. Finally, the concept without the sharing unit could be extended to load-bearing legged robots in general, using a centralized adjustable static force unit for all legs.

\section*{Acknowledgments}
This work was supported by the Fonds québécois de la recherche sur la nature et les technologies (FRQNT) [DOI 285727 and 303608]; and the Natural Sciences and Engineering Research Council of Canada (NSERC).

\bibliographystyle{elsarticle-num} 
\bibliography{references}

\end{document}